# From Angular Manifolds to the Integer Lattice: Guaranteed Orientation Estimation with Application to Pose Graph Optimization

Luca Carlone        Andrea Censi

*Abstract*—Estimating the orientations of nodes in a pose graph from relative angular measurements is challenging because the variables live on a manifold product with nontrivial topology and the maximum-likelihood objective function is non-convex and has multiple local minima; these issues prevent iterative solvers to be robust for large amounts of noise. This paper presents an approach that allows working around the problem of multiple minima, and is based on the insight that the original estimation problem on orientations is equivalent to an unconstrained quadratic optimization problem on integer vectors. This equivalence provides a viable way to compute the maximum likelihood estimate and allows guaranteeing that such estimate is almost surely unique. A deeper consequence of the derivation is that the maximum likelihood solution does not necessarily lead to an estimate that is "close" to the actual nodes orientations, hence it is not necessarily the best choice for the problem at hand. To alleviate this issue, our algorithm computes a *set* of estimates, for which we can derive precise probabilistic guarantees. Experiments show that the method is able to tolerate extreme amounts of noise (e.g., $\sigma = 30°$ on each measurement) that are above all noise levels of sensors commonly used in mapping. For most range-finder-based scenarios, the multi-hypothesis estimator returns only a single hypothesis, because the problem is very well constrained. Finally, using the orientations estimate provided by our method to bootstrap the initial guess of pose graph optimization methods improves their robustness and makes them avoid local minima even for high levels of noise.

## I. INTRODUCTION

A *pose graph* is a model used in probabilistic robotics to formalize the Simultaneous Localization and Mapping (SLAM) problem [1]. Each node in the graph represents the pose of a mobile robot at a given time, whereas an edge exists between two nodes if a relative measurement (inter-nodal constraint) is available between the two poses. Relative measurements might be obtained by means of proprioceptive sensors (e.g., wheel odometry) or exteroceptive-sensor-based techniques (e.g., scan matching or visual odometry).

The objective of *pose graph optimization* is to estimate the nodes poses that maximize the likelihood of inter-nodal measurements. Once the poses have been estimated, it is possible to construct a map of the environment by placing all measurements in the same global coordinate frame.

The difficulty in obtaining the maximum likelihood estimate of robot poses is mainly connected with the angular component: also in a planar case, the nodes orientations belong to a product of manifolds ($SO(2)^n$, with $n$ the number of observable poses) that have a nontrivial topology. This makes the maximum likelihood problem nonlinear and non-convex,

L. Carlone (*corresponding author*) is with the Dipartimento di Automatica e Informatica, Politecnico di Torino, Torino, Italy. luca.carlone@polito.it
A. Censi is with the Control & Dynamical Systems department, California Institute of Technology, Pasadena, CA, USA. andrea@cds.caltech.edu

with multiple local minima (see Wang *et al.* [2] for an exhaustive analysis of a simple instance of the problem). In fact, if the orientations were known, pose optimization would be a linear problem, see e.g., [3].

This paper considers the problem of estimating the nodes orientation from pairwise relative angular measurements, which is referred to as the *orientation graph optimization* problem. We provide a multi-hypothesis global optimization method that does not suffer from local minima, even with extreme amounts of noise. In the context of SLAM, we will show that having a global estimate of the orientations improves the robustness of iterative solvers such as g2o [4].

*Related work in robotics*: The formulation of SLAM as a nonlinear optimization problem on a graph traces back to Lu and Milios [5]. Gutmann and Konolige [6] discuss how to build a pose graph in incremental fashion from laser scan measurements. A large amount of subsequent work focuses on speeding up computation. Duckett *et al.* [7] use a Gauss-Seidel relaxation to minimize residual errors. Konolige [8] describes a reduction scheme to improve efficiency of non-linear optimization. Thrun and Montemerlo [1] describe a conjugate gradient-based optimization that enables large scale estimation. Frese *et al.* [9] propose a multilevel relaxation technique that considerably reduces the computation time by applying a multi-grid algorithm. Olson *et al.* [10] propose an alternative parametrization for the problem, which entails several advantages in terms of computation and robustness. Grisetti *et al.* [11] extend such framework, proposing a method (Toro) that is based on stochastic gradient descent and uses a tree-based parametrization to optimize the poses in both planar and three-dimensional scenarios. Kaess *et al.* [12]–[14] present a very elegant formalization of SLAM using a *Bayes tree* model and investigate incremental estimation techniques. Several recent papers focus on the manifold structure of the problem: the domain of the problem is a product of manifolds SE(2) or SE(3), and this aspect requires a suitable treatment when using iterative optimization techniques [15], or closed-form problem-specific methods [16]–[18]. Kuemmerle *et al.* [4] describe the g2o framework for solving general optimization problems with variables belonging to manifolds. Olson and Agarwal [19], and Sünderhauf and Protzel [20], [21] propose relevant extensions of this framework, with the purpose of increasing estimation robustness in the presence of outliers. The theoretical analysis of the problem is slightly behind applications; see Knuth and Barooah [22], Huang *et al.* [23], and the previously mentioned Wang *et al.* [2].

The state-of-the-art techniques for pose graph optimization are iterative approaches that minimize a cost function starting from an initial guess. None can guarantee convergence to a global minimum, and it is observed that they get easily trapped in local minima in presence of large orientation noise.

*Related work in other fields*: In this paper we limit ourselves to the robotics perspective of pose graph optimization and relative benchmarks, but one must point out that there exist many other applications, such as attitude synchronization [24] and calibration of camera networks [25], which consider problems that are formally equivalent or very similar to pose graph optimization. It is common for these problems to be formulated in a multi-agent context, where the problem is to estimate in a distributed way some local state of the agent (pose, position, orientation, etc.) with many variations according to the kind of measurements available (relative distance, relative bearing, etc.). For example, Barooah and Hespanha [3], [26] consider the problem of estimating positions of robots in a team from relative position measurements, assuming known orientations. Knuth and Barooah focuses on distributed computation [27]. The case in which the nodes positions have to be estimated from bearing measurements was pioneered by Stanfield [28] and further developed in more recent work [29]–[31]. Another common setup is the one in which nodes positions are estimated from pairwise distance measurements [32]–[35].

*Paper outline*: Our results derive from the joint application of graph theory, differential geometry, and integer programming. We do not assume any prior knowledge and Section II recalls all necessary preliminaries.

Section III recalls the usual maximum likelihood formalization for the orientation estimation problem, with extra care to the assumptions and the problem symmetries.

Section IV proves that the maximum likelihood optimization problem with domain $\mathrm{SO}(2)^n$, where $n$ is the number of observable nodes, is equivalent to an unconstrained quadratic integer optimization problem on $\mathbb{Z}^\ell$, where $\ell$ is the number of cycles in the graph (Theorem 16). First, we show that it is possible to map the nonlinear maximum likelihood estimation problem from the manifold to a vector space, by including integer-valued unknowns (*regularization terms*). The corresponding maximum likelihood problem becomes a mixed-integer program [36]. Further, the objective function for this problem can be separated into two terms in a way that allows a two-stage optimization, in which one first optimizes over the regularization terms, and then the maximum likelihood estimate of the nodes orientation is computed in a closed form. The conclusion, given in Theorem 16, is that the maximum likelihood estimate is unique with probability one, and that the global maximum of the likelihood function can be found by solving an unconstrained quadratic integer program.

Section V describes several properties of the probability distribution of the maximum likelihood estimator that imply that the estimate may suffer from a bias. As a simple example demonstrates, since the problem is highly nonlinear, minimizing the *data error* does not imply that the *estimation error* is low. By contrast, in a linear problem, the maximum likelihood estimator is also unbiased and a minimum variance estimator. Our conclusion is that in this problem the maximum likelihood estimate is not necessarily the most useful information when the noise is large. Motivated by this result we look for a multi-hypothesis estimator for which we can derive stronger results.

Section VI and Section VII describe the MOLE2D algorithm, which returns a set of multiple hypotheses for the nodes orientations. We can give probabilistic guarantees on the output of this algorithm, namely that at least one hypothesis is "close" to the actual nodes orientations within a given confidence level. The algorithm is able to return only a small set of plausible hypotheses, provided that the "frame of reference" is chosen appropriately. The frame of reference is given by the free choice of a *cycle basis matrix* that must be supplied. It can be proven that choosing the *minimum* cycle basis minimizes the expected number of hypotheses. However, because the exact minimum cycle basis is expensive to compute, it is worth exploring several approximations as alternatives. With the right choice of cycle basis, in common problem instances the set of estimates contains a single element, because the problem is very well constrained. In this case, we are able to completely characterize the distribution of the estimator, which is rare in the context of nonlinear estimation.

Section VIII discusses the performance of MOLE2D on standard SLAM datasets, both for orientation estimation and for full pose optimization. For the case of orientation estimation, we explore the trade-off in performance implied by the choice of the cycle basis matrix used by MOLE2D. The results confirm the theoretical predictions. For the case of pose optimization, we show that simply substituting the orientation estimate computed by MOLE2D in place of the odometric initial guess greatly enhances the robustness to noise in an iterative solver such as g2o.

*Relation with previous work*: Previous work by the first author and colleagues [37], [38] highlighted the importance of orientation estimation in pose graph optimization and proposed a fast approximation for solving for the orientations, and then solving for the translations given the orientations. Such previous work motivated this development, but the approach of the present work follows a different route. This paper presents a formal treatment of the orientation-only estimation problem; rather than proposing an approximation, we care about finding the *exact* maximum likelihood (orientation) estimate. In hindsight, the results of this paper allow to conclude that the rounding operation proposed to solve the wraparound problem in [38] is only a heuristic to solve a quadratic integer program, which does not necessarily lead to the optimal solution [39]. This paper also asserts in Section V that, after all, the maximum likelihood estimate may not be the best choice for the problem at hand, hence it proposes to switch to a multi-hypothesis approach. More in detail, the proof of equivalence to a quadratic integer program, the MOLE2D algorithm, and the experimental results are original and have not been published in previous work or submitted to conferences.

## II. PRELIMINARIES

This section introduces some preliminaries of graph theory, differential geometry, and modulus algebra. Table I summarizes the most important symbols appearing in the paper.

TABLE I
SYMBOLS USED IN THIS PAPER

*Graph*

| | |
|---|---|
| $\mathcal{G} = (\mathcal{V}, \mathcal{E})$ | Directed graph |
| $m$ | Number of edges |
| $n+1$ | Number of nodes |
| $n$ | Number of observable variables |
| $\mathcal{V}$ | Vertex set; $|\mathcal{V}| = n+1$ |
| $\mathcal{E}$ | Edge set; $|\mathcal{E}| = m$ |
| $e = (i,j) \in \mathcal{E}$ | Edge between nodes $i$ and $j$ |
| $\ell$ | Number of cycles; $\ell = m - n$ |
| $w_{ij}$ | Weight associated to edge $(i,j)$ |
| $\overline{\boldsymbol{A}} \in \mathbb{R}^{(n+1) \times m}$ | Incidence matrix of $\mathcal{G}$ |
| $\boldsymbol{A} \in \mathbb{R}^{n \times m}$ | Reduced incidence matrix of $\mathcal{G}$ |

*Cycle bases*

| | |
|---|---|
| $\mathcal{C}_\mathcal{G}$ | The set of all cycle basis for $\mathcal{G}$ |
| $\boldsymbol{c} \in \{-1, 0, +1\}^m$ | Row vector describing a circuit |
| $W(\boldsymbol{c}, \boldsymbol{w})$ | Weight of a cycle |
| $W(\boldsymbol{C}, \boldsymbol{w})$ | Weight of a cycle basis |
| $\text{MCB}(\mathcal{G}, \boldsymbol{w})$ | Minimum cycle basis for a given weight function |
| $\boldsymbol{C} \in \mathbb{Z}^{\ell \times m}$ | Matrix describing a cycle basis |
| $\boldsymbol{C}_L, \boldsymbol{C}_T$ | Canonical ordering of $\boldsymbol{C}$ according to a spanning tree $T$ |
| FCB | Fundamental cycle basis built from a spanning tree |

*Geometry of angles*

| | |
|---|---|
| $\text{SO}(2)$ | 2D rotation matrices |
| $\text{Exp}: \mathbb{R} \to \text{SO}(2)$ | Exponential map |
| $\text{Log}: \text{SO}(2) \to \mathcal{P}(\mathbb{R})$ | Logarithmic map |
| $\text{Log}_0: \text{SO}(2) \to \mathbb{R}$ | Principal logarithmic map |
| $\langle \cdot \rangle_{2\pi}: \mathbb{R} \to (-\pi, +\pi]$ | $2\pi$ modulus operation |

*Orientation estimation (intrinsic formalization)*

| | |
|---|---|
| $\boldsymbol{r}_i^\circ \in \text{SO}(2)$ | Unknown node orientation |
| $\boldsymbol{r}_i \in \text{SO}(2)$ | Optimization variables for node orientation |
| $\boldsymbol{d}_{ij} \in \text{SO}(2)$ | Relative orientation measurement |
| $\boldsymbol{\varepsilon}_{ij} \in \text{SO}(2)$ | Measurement error |
| $\epsilon_{ij} \in \mathbb{R}$ | Gaussian noise producing $\boldsymbol{\varepsilon}_{ij}$ |
| $\sigma_{ij}$ | Standard deviation of $\epsilon_{ij}$ |

*Orientation estimation (in $(-\pi, +\pi]$ coordinates)*

| | |
|---|---|
| $\check{\boldsymbol{\theta}}^\circ \in (-\pi, +\pi]^n$ | Unknown orientations |
| $\check{\boldsymbol{\theta}} \in (-\pi, +\pi]^n$ | Optimization variables for orientations |
| $\check{\boldsymbol{\delta}} \in (-\pi, +\pi]^m$ | Relative orientation measurements |
| $\boldsymbol{P}_\delta \in \mathbb{R}^{m \times m}$ | Measurement covariance |

*Mixed-integer formalization in $\boldsymbol{k}$ and $\boldsymbol{\theta}$*

| | |
|---|---|
| $\boldsymbol{\theta} \in \mathbb{R}^n$ | Real-valued optimization variables for orientations |
| $\boldsymbol{k} \in \mathbb{Z}^m$ | Regularization vector |
| $\check{\boldsymbol{\theta}}^{\star\|\boldsymbol{k}}$ | Estimate of $\check{\boldsymbol{\theta}}^\circ$ given $\boldsymbol{k} \in \mathbb{Z}^m$ |

*Reduced formalization in cycle space*

| | |
|---|---|
| $\boldsymbol{\gamma} \in \mathbb{Z}^\ell$ | Integer vector living on the cycles |
| $\hat{\boldsymbol{\gamma}} \in \mathbb{Z}^m$ | Estimator for $\boldsymbol{\gamma}$ |
| $\boldsymbol{\theta}^{\star\|\boldsymbol{\gamma}}$ | Real-valued estimate of $\check{\boldsymbol{\theta}}^\circ$ given $\boldsymbol{\gamma}$ |
| $\check{\boldsymbol{\theta}}^{\star\|\boldsymbol{\gamma}} = \langle \boldsymbol{\theta}^{\star\|\boldsymbol{\gamma}} \rangle_{2\pi}$ | Wrapped estimate of $\check{\boldsymbol{\theta}}^\circ$ given $\boldsymbol{\gamma}$ |
| $\check{\boldsymbol{\theta}}^\star = \check{\boldsymbol{\theta}}^{\star\|\boldsymbol{\gamma}^\star}$ | Max. likelihood estimate of $\check{\boldsymbol{\theta}}^\circ$ |
| $\Gamma$ | Set of estimates for $\boldsymbol{\gamma}$ returned by INTEGER-SCREENING |
| $\Theta$ | Set of estimates of $\check{\boldsymbol{\theta}}^\circ$ returned by MOLE2D |

*Miscellanea*

| | |
|---|---|
| $\mathcal{P}(S)$ | Power set of the set $S$ |
| $|S|$ | Cardinality of the set $S$ |
| $\mathbf{I}_n$ | $n \times n$ identity matrix |
| $\mathbf{0}_n$ | (column) Vector of all zeros of dimension $n$ |
| $\mathbf{1}_n$ | (column) Vector of all ones of dimension $n$ |
| $\lfloor \cdot \rfloor$ | Floor operator |
| $\text{Trace}\{\boldsymbol{P}\}$ | Trace of the matrix $\boldsymbol{P}$ |
| $\chi^2_{\ell, \alpha}$ | quantile of the $\chi^2$ distribution with $\ell$ degrees of freedom and upper tail probability equal to $\alpha$ |

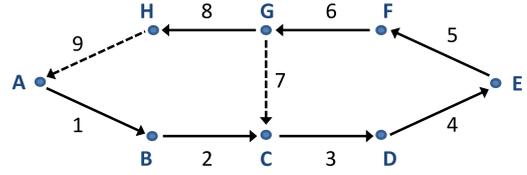

$$\overline{\boldsymbol{A}} = \begin{pmatrix} -1 & 0 & 0 & 0 & 0 & 0 & 0 & 0 & +1 \\ +1 & -1 & 0 & 0 & 0 & 0 & 0 & 0 & 0 \\ 0 & +1 & -1 & 0 & 0 & 0 & +1 & 0 & 0 \\ 0 & 0 & +1 & -1 & 0 & 0 & 0 & 0 & 0 \\ 0 & 0 & 0 & +1 & -1 & 0 & 0 & 0 & 0 \\ 0 & 0 & 0 & 0 & +1 & -1 & 0 & 0 & 0 \\ 0 & 0 & 0 & 0 & 0 & +1 & -1 & -1 & 0 \\ 0 & 0 & 0 & 0 & 0 & 0 & 0 & +1 & -1 \end{pmatrix}$$

$$\boldsymbol{C}_1 = \begin{pmatrix} +1 & +1 & +1 & +1 & +1 & +1 & 0 & +1 & +1 \\ +1 & +1 & 0 & 0 & 0 & 0 & -1 & +1 & +1 \end{pmatrix}$$

$$\boldsymbol{C}_2 = \begin{pmatrix} +1 & +1 & 0 & 0 & 0 & 0 & -1 & +1 & +1 \\ 0 & 0 & +1 & +1 & +1 & +1 & +1 & 0 & 0 \end{pmatrix}$$

Fig. 1. A toy graph with vertex set $\mathcal{V}$ comprising nodes from $A$ to $H$, and edge set $\mathcal{E}$ comprising edges 1 to 9. For our example we assume that each edge has a unitary weight. A spanning tree (which is also a spanning *path* in this case) is given by edges $\{1,2,3,4,5,6,8\}$; the corresponding chords are edge 7 and edge 9 (reported as dashed lines in the figure). The figure also shows the incidence matrix of the graph, while the reduced incidence matrix $\boldsymbol{A}$ is obtained from $\overline{\boldsymbol{A}}$ by deleting the first row. $\boldsymbol{C}_1$ is a cycle basis matrix for the graph, whose first circuit includes edges $\{1,2,3,4,5,6,8,9\}$ and second circuit includes edges $\{3,4,5,6,7\}$. $\boldsymbol{C}_2$ is a minimum cycle basis, whose first circuit includes edges $\{3,4,5,6,7\}$ and second circuit includes edges $\{1,2,7,8,9\}$.

### A. Computational graph theory

Chen [40] is a popular reference for standard concepts of computational graph theory. Our notation is compatible with Kavitha *et al.* [41], from which we take the more specific results about cycle bases.

A *directed graph* $\mathcal{G}$ is a pair $(\mathcal{V}, \mathcal{E})$, where $\mathcal{V}$ is a finite set of elements, called *vertices* or *nodes*, and $\mathcal{E}$ is a set containing ordered pairs of nodes. A generic element $e \in \mathcal{E}$, referred to as *edge*, is in the form $e = (i, j)$, meaning that edge $e$, incident on nodes $i$ and $j$, leaves node $i$ and is directed towards node $j$ ($i$ is called *tail* and $j$ is called *head*).

A *weighted graph* has also a nonnegative weight associated to each edge; we denote with $\boldsymbol{w}$ the weight function that associates a weight $w_{ij}$ to each edge $e = (i, j)$.

The number of nodes and edges are denoted with $n+1$ and $m$, respectively, i.e., $|\mathcal{V}| = n+1$ and $|\mathcal{E}| = m$. Our graph has $n+1$ nodes, rather than $n$, because only $n$ independent variables will be observable, and this choice will simplify the notation later on.

The *incidence matrix* $\overline{\boldsymbol{A}}$ of a directed graph is a matrix in $\{-1, 0, +1\}^{(n+1) \times m}$ that describes the structure of the graph. Each column of $\overline{\boldsymbol{A}}$ corresponds to an edge, and the column corresponding to edge $e = (i, j)$ has only two non-zero elements, one on the $i$-th row (equal to $-1$) and the other on the $j$-th row (equal to $+1$). Figure 1 shows an intuitive example.

The *reduced incidence matrix* $\boldsymbol{A}$ is obtained from $\overline{\boldsymbol{A}}$ by removing one row. Without loss of generality, in this paper we assume that it is the first row, which corresponds to the first node that is set to the origin of the reference frame. If $\overline{\boldsymbol{A}}$ has dimensions $n+1 \times m$, $\boldsymbol{A}$ has dimension $n \times m$.

A *spanning tree* of a graph is a subgraph with $n$ edges that contains all the nodes in the graphs. For a given spanning tree, the edges of the original graph that do not belong to the spanning tree are called *chords*.

A *cycle* is a subgraph in which every node appears in an even number of edges. A *circuit* is a cycle in which every node appears exactly in two edges. A (directed) circuit can be described by a vector of $m$ elements in which the $e$-th element is $+1$ or $-1$ if edge $e$ is traversed respectively forwards (from tail to head) or backwards, and $0$ if it does not appear in the circuit. Therefore, a circuit can be represented by a vector in $\{-1, 0, +1\}^m$. In a cycle, instead, an edge can appear twice or more. Correspondingly, a cycle is represented by a vector $\boldsymbol{c} \in \mathbb{Z}^m$. If the graph is weighted, then we can associate a weight to any cycle, by summing together the weights of the edges traversed by the cycle. If a cycle is described by a vector $\boldsymbol{c} \in \mathbb{Z}^m$, then the weight of the cycle is given by

$$\boldsymbol{W}(\boldsymbol{c}, \boldsymbol{w}) = \sum_{(i,j) \in \mathcal{E}} w_{ij} |c_{ij}|. \tag{1}$$

A *cycle basis* of a graph is a minimal set of circuits such that any cycle in the graph can be written as a combination of the circuits in the basis. We define $\mathcal{C}_\mathcal{G}$ as the set of all cycle basis of the graph. The number of independent circuits in the cycle basis is called *cyclomatic number* and it is equal to $\ell = m - n$.

A *cycle basis matrix* is a matrix $\boldsymbol{C} \in \mathbb{Z}^{\ell \times m}$, such that each row $\boldsymbol{c}^{(t)}$ describes one of the circuits in the cycle basis:

$$\boldsymbol{C} = \begin{pmatrix} \boldsymbol{c}^{(1)} \\ \vdots \\ \boldsymbol{c}^{(t)} \\ \vdots \\ \boldsymbol{c}^{(\ell)} \end{pmatrix} \in \mathbb{Z}^{\ell \times m}. \tag{2}$$

The weight of a cycle basis is the sum of the cycles weights:

$$\boldsymbol{W}(\boldsymbol{C}, \boldsymbol{w}) = \sum_{t=1}^{\ell} \boldsymbol{W}(\boldsymbol{c}^{(t)}, \boldsymbol{w}). \tag{3}$$

If there is a weight associated to every cycle basis, then we can look for a basis that has minimum weight. This is called *minimum cycle basis* (MCB):

$$\mathrm{MCB}(\mathcal{G}, \boldsymbol{w}) \doteq \arg\min_{\boldsymbol{C} \in \mathcal{C}_\mathcal{G}} \boldsymbol{W}(\boldsymbol{C}, \boldsymbol{w}). \tag{4}$$

In the rest of the paper we consider a weight function $\boldsymbol{w}$ that associates to each edge the variance of the corresponding measurement, as formalized in Section III. Therefore, we use the notation "MCB" omitting the dependence on the graph and on the weight function, and implying that we consider a minimum *uncertainty* cycle basis.

Cycle bases matrices and incidence matrices have an array of interesting properties.

**Lemma 1** (Orthogonal complements [42]). *For a connected graph $\mathcal{G}$, the transpose of the cycle basis matrix $\boldsymbol{C}^\mathsf{T}$ is an orthogonal complement of the transpose of the reduced incidence matrix $\boldsymbol{A}^\mathsf{T}$, i.e.,*

1) $(\boldsymbol{A}^\mathsf{T} \; \boldsymbol{C}^\mathsf{T})$ *is a square matrix of full rank; and*
2) $\boldsymbol{C}\boldsymbol{A}^\mathsf{T} = \boldsymbol{0}_{\ell \times n}$.

In order to simplify the notation, and without loss of generality, we order the edges such that the first $n$ edges belong to a spanning tree $T$ and the remaining $\ell$ edges are chords with respect to $T$. This allows to write the cycle basis matrix $\boldsymbol{C}$ as

$$\boldsymbol{C} = (\boldsymbol{C}_T \; \boldsymbol{C}_L), \tag{5}$$

where $\boldsymbol{C}_T \in \mathbb{Z}^{\ell \times n}$ contains the columns in $\boldsymbol{C}$ corresponding to edges in $T$, and $\boldsymbol{C}_L \in \mathbb{Z}^{\ell \times \ell}$ contains the columns in $\boldsymbol{C}$ corresponding to chords with respect to $T$.

### B. Modulus operation

The map $\langle \cdot \rangle_{2\pi}$ is a function from $\mathbb{R}$ to the interval $(-\pi, +\pi]$:

$$\langle \cdot \rangle_{2\pi} : \mathbb{R} \to (-\pi, +\pi], \tag{6}$$

which can be written explicitly as

$$\langle \omega \rangle_{2\pi} \doteq \omega + 2\pi \left\lfloor \frac{\pi - \omega}{2\pi} \right\rfloor \in (-\pi, +\pi], \tag{7}$$

where $\lfloor \cdot \rfloor$ is the floor operator. Therefore, for a given $\omega$, it is well defined the value $k_\omega \in \mathbb{Z}$ such that

$$\langle \omega \rangle_{2\pi} = \omega + 2\pi k_\omega. \tag{8}$$

The integer $k_\omega = \lfloor \frac{\pi - \omega}{2\pi} \rfloor$ is the regularization term necessary for the result to be in $(-\pi, +\pi]$ [43].

Notice that for all $\hat{k} \in \mathbb{Z}$, it holds $|\omega + 2\pi k_\omega| \leq |\omega + 2\pi \hat{k}|$, since $\langle \omega \rangle_{2\pi} = \omega + 2\pi k_\omega \in (-\pi, +\pi]$ (by definition of the map $\langle \cdot \rangle_{2\pi}$) and we cannot further reduce the absolute value of a quantity in $(-\pi, +\pi]$ by adding or subtracting a multiple of $2\pi$. Therefore, it also holds that

$$k_\omega = \arg\min_{k \in \mathbb{Z}} |\omega + 2\pi k|, \tag{9}$$

and, for the same reasoning,

$$|\langle \omega \rangle_{2\pi}| = |\omega + 2\pi k_\omega| = \min_{k \in \mathbb{Z}} |\omega + 2\pi k|. \tag{10}$$

This is a simple property of $\langle \cdot \rangle_{2\pi}$:

$$\langle \omega_1 + \omega_2 \rangle_{2\pi} = \langle \langle \omega_1 \rangle_{2\pi} + \langle \omega_2 \rangle_{2\pi} \rangle_{2\pi}. \tag{11}$$

### C. Differential geometry of angles

The *exponential map* for the manifold $\mathrm{SO}(2)$ is a map from the *tangent space* $\mathrm{so}(2) \simeq \mathbb{R}$ to the manifold:

$$\mathrm{Exp} : \mathbb{R} \to \mathrm{SO}(2). \tag{12}$$

This map is onto (surjective) but not 1-to-1 (bijective).

The *logarithmic map* is the *right inverse* of the exponential map, and it maps an angle in $\mathrm{SO}(2)$ to all possible elements in the tangent space that have the same exponential:

$$\mathrm{Log} : \mathrm{SO}(2) \to \mathcal{P}(\mathbb{R}). \tag{13}$$

Here, "$\mathcal{P}(\mathbb{R})$" denotes the *power set* of $\mathbb{R}$. Note that the fact that the exponential map is not invertible is an intrinsic property and does not depend on the choice of a particular

parametrization. The logarithmic map satisfies the Lie group property

$$\text{Log}(\boldsymbol{s}^{-1}) = -\text{Log}(\boldsymbol{s}) \quad (14)$$

and the Abelian property

$$\text{Log}(\boldsymbol{s}_1 \boldsymbol{s}_2) = \text{Log}(\boldsymbol{s}_1) + \text{Log}(\boldsymbol{s}_2). \quad (15)$$

The *principal logarithm map* $\text{Log}_0$ is a 1-to-1 function that chooses one particular element on the tangent space:

$$\text{Log}_0 : \text{SO}(2) \to \mathbb{R}, \quad (16)$$

namely, the closest to the origin. A property of the principal logarithm is that

$$\text{Log}_0(\boldsymbol{s}) = \langle \text{Log}(\boldsymbol{s}) \rangle_{2\pi}. \quad (17)$$

If we parametrize the manifold $\text{SO}(2)$ with angular coordinates in $(-\pi, +\pi]$, then the coordinate version of $\text{Exp}$ is simply the modulus $\langle \cdot \rangle_{2\pi}$, while the principal logarithm maps a rotation matrix to the corresponding angle of rotation in $(-\pi, +\pi]$.

### D. Wrapped Gaussian distribution on the circle

The *wrapped* Gaussian distribution on the circle is the generalization of a Gaussian distribution [44], [45], in the sense that it is the solution of the heat equation on the circle, and has several other analogous properties. It can be obtained by applying the exponential map to a Gaussian variable that lives on the tangent space:

$$\boldsymbol{\varepsilon} = \text{Exp}(\epsilon), \quad \text{with } \epsilon \sim \mathcal{N}(0, \sigma^2). \quad (18)$$

The probability density function for a wrapped Gaussian $\mathcal{W}_{\sigma^2} : \text{SO}(2) \to \mathbb{R}^+$ can be written as

$$\mathcal{W}_{\sigma^2}(\boldsymbol{\varepsilon}) = \frac{1}{\sigma\sqrt{2\pi}} \sum_{k=-\infty}^{+\infty} \exp\left(\frac{-(\text{Log}_0(\boldsymbol{\varepsilon}) + 2\pi k)^2}{2\sigma^2}\right). \quad (19)$$

Note that $\text{Log}_0$ returns a value in $[-\pi, +\pi)$.

A wrapped Gaussian may show a very different behaviour w.r.t. a Gaussian density. For instance, as the noise increases, a Gaussian density would tend pointwise to 0, while the wrapped Gaussian distribution tends to the uniform distribution on the circle:

$$\lim_{\sigma^2 \to \infty} \mathcal{W}_{\sigma^2} = \tfrac{1}{2\pi}. \quad (20)$$

Other properties are instead maintained, such as the closure with respect to convolution [45].

**Lemma 2.** *If $\boldsymbol{s}_1 \sim \mathcal{W}_{\sigma_1^2}$ and $\boldsymbol{s}_2 \sim \mathcal{W}_{\sigma_2^2}$, then the product $\boldsymbol{s}_1 \boldsymbol{s}_2$ has distribution $\mathcal{W}_{\sigma_1^2 + \sigma_2^2}$.*

### III. PROBLEM STATEMENT: MAXIMUM LIKELIHOOD ORIENTATION ESTIMATION

Let $\mathcal{G}$ be a directed graph with $n+1$ nodes and $m$ edges, and call $\mathcal{E}$ the set of edges in $\mathcal{G}$. Let each node be assigned an unknown orientation represented by a rotation matrix $\boldsymbol{r}_i^\circ \in \text{SO}(2)$. Suppose that it is possible to measure the relative orientation of two nodes sharing an edge. For any edge $(i,j) \in \mathcal{E}$, the observation $\boldsymbol{d}_{ij} \in \text{SO}(2)$ is a noisy measurement of the relative orientation:

$$\boldsymbol{d}_{ij} = (\boldsymbol{r}_i^\circ)^{-1} \boldsymbol{r}_j^\circ \, \boldsymbol{\varepsilon}_{ij} \in \text{SO}(2), \quad (21)$$

where $\boldsymbol{r}_i^\circ$ is the true (unknown) orientation of the $i$-th node and $\boldsymbol{\varepsilon}_{ij}$ is a random variable on $\text{SO}(2)$ that represents the noise in the measurements, and that we assume to be distributed according to a *wrapped* Gaussian with variance $\sigma_{ij} > 0$ (see Section II-D).

The graph $\mathcal{G}$ is *weighted*, and the corresponding weight function is $w : (i,j) \to \sigma_{ij}^2$, so that the weight of an edge is set to the variance of the corresponding orientation measurement.

For now, all quantities are members of the manifold $\text{SO}(2)$. This first formalization of the orientation estimation problem is thus intrinsic on the manifold.

**Problem 1** (Intrinsic formulation of maximum-likelihood orientation estimation in the absolute frame)**.** *Given the set of relative observations $\{\boldsymbol{d}_{ij}\} \in \text{SO}(2)^m$, for $(i,j) \in \mathcal{E}$ and the corresponding variances $\sigma_{ij} > 0$, find the set of minimizers $S^1 \subset \text{SO}(2)^{n+1}$ that satisfies*

$$S^1 = \operatorname*{arg\,min}_{\{\boldsymbol{r}_i\} \in \text{SO}(2)^{n+1}} \sum_{(i,j) \in \mathcal{E}} -\log \mathcal{W}_{\sigma_{ij}^2}(\boldsymbol{d}_{ij}^{-1} \boldsymbol{r}_i^{-1} \boldsymbol{r}_j). \quad (22)$$

We are using the symbol $\boldsymbol{r}_i$ to denote the optimization variable associated to the orientation of the $i$-th node, while $\boldsymbol{r}_i^\circ$ is the true orientation of the node. Adding Gaussian priors is easy by using virtual measurements, but we do not do it explicitly.

In this paper, we pose all optimization problems as finding a *set* of minimizers, rather than the "optimal solution". The set of minimizers is indicated as $S^i$ for Problem $i$. Only in some cases we will be able to conclude there is a unique solution, and hence $S^i$ has only one element. We need to be careful about keeping track of the set of minimizers, because the first part of the paper (Section IV) consists in transforming one problem to another, sometimes changing the domain or introducig extra variables. To facilitate bookkeeping, we use the concept of *symmetry*: a symmetry of an optimization problem is an invertible transformation of the unknown variables that preserves the value of the objective function. Speaking of symmetries is a formal way of speaking of unobservability from the algebraic/geometric point of view.

Problem 1 has one symmetry that corresponds to the well-known fact that the absolute orientations are not observable from only relative measurements: the relative measurements do not change if the nodes orientations are rotated by the same amount. Formally, for any rotation matrix $\boldsymbol{s}$, the objective function (22) is invariant if we apply the invertible function

$$f_{\boldsymbol{s}} : \quad \text{SO}(2)^{n+1} \quad \to \text{SO}(2)^{n+1} \quad (23)$$
$$\{\boldsymbol{r}_i\} \quad \mapsto \{\boldsymbol{s}\, \boldsymbol{r}_i\}. \quad (24)$$

Following standard procedures, to avoid this ambiguity we fix the orientation of the first node to the arbitrary value $\boldsymbol{r}_0 = \begin{pmatrix} 1 & 0 \\ 0 & 1 \end{pmatrix}$. This corresponds to setting the absolute frame aligned with the first robot pose. Therefore, we can restate the problem considering only $n$ nodes instead of $n+1$.

TABLE II
RELATIONS AMONG PROBLEMS DEFINED IN THIS PAPER

| problem | variables | solutions | symmetries of solutions set |
|---|---|---|---|
| Problem 1 | $\{r_i\} \in \mathrm{SO}(2)^{n+1}$ | $S^1$ | For any $s \in \mathrm{SO}(2)$, $\{r_i\} \mapsto \{sr_i\}$. |
| $\downarrow$ Fixing reference frame | | | |
| Problem 2 | $\{r_i\} \in \mathrm{SO}(2)^n$ | $S^2$ | none |
| $\downarrow$ Choice of coordinates | | $\mathrm{Log}_0 \downarrow \uparrow \mathrm{Exp}$ | |
| Problem 3 | $\check{\boldsymbol{\theta}} \in (-\pi,+\pi]^n$ | $S^3$ | none |
| $\downarrow$ Real-valued parametrization | | $\varphi_4^3 \uparrow$ | |
| Problem 4 | $\boldsymbol{\theta} \in \mathbb{R}^n$ | $S^4$ | For any $\boldsymbol{p} \in \mathbb{Z}^n$, $\boldsymbol{\theta} \mapsto \boldsymbol{\theta} - 2\pi\boldsymbol{p}$. |
| $\downarrow$ Introduction of regularization terms | | $\varphi_5^4 \uparrow$ | |
| Problem 5 | $(\boldsymbol{\theta}, \boldsymbol{k}) \in \mathbb{R}^n \times \mathbb{Z}^m$ | $S^5$ | For any $\boldsymbol{p} \in \mathbb{Z}^n$, $(\boldsymbol{\theta}, \boldsymbol{k}) \mapsto (\boldsymbol{\theta} - 2\pi\boldsymbol{p}, \boldsymbol{k} + \boldsymbol{A}^\mathsf{T}\boldsymbol{p})$. |
| $\downarrow$ Separability of error function | | $\varphi_6^5 \uparrow$ | |
| Problem 6 | $\boldsymbol{k} \in \mathbb{Z}^m$ | $S^6$ | For any $\boldsymbol{p} \in \mathbb{Z}^n$, $\boldsymbol{k} \mapsto \boldsymbol{k} + \boldsymbol{A}^\mathsf{T}\boldsymbol{p}$. |
| $\downarrow$ Minimality of parametrization | | $\varphi_7^6 \uparrow$ | |
| Problem 7 | $\boldsymbol{\gamma} \in \mathbb{Z}^\ell$ | $S^7$ | none |

**Problem 2.** *(Intrinsic formulation of maximum-likelihood orientation estimation)* Given a set of relative observations $\boldsymbol{d}_{ij} \in \mathrm{SO}(2)^m$, for $(i,j) \in \mathcal{E}$, and the corresponding variances $\sigma_{ij} > 0$, find the set of minimizers $S^2 \subset \mathrm{SO}(2)^n$ that satisfies

$$S^2 = \underset{\{r_i\} \in \mathrm{SO}(2)^n}{\arg\min} \sum_{(i,j) \in \mathcal{E}} -\log \mathcal{W}_{\sigma_{ij}^2}(\boldsymbol{d}_{ij}^{-1} \boldsymbol{r}_i^{-1} \boldsymbol{r}_j), \quad (25)$$

having fixed $\boldsymbol{r}_0 = \begin{pmatrix} 1 & 0 \\ 0 & 1 \end{pmatrix}$.

It is easy to see that the function must admit a minimum, as it is bounded below and defined on the compact set $\mathrm{SO}(2)^n$.

Does this problem admit a unique minimum? This is certainly the case in a noiseless setup and for a connected graph, as proven by Proposition 3.

**Proposition 3.** *(Observability of orientations)* In the noiseless case, $S^2$ contains exactly one element if and only if the graph is connected.

*Proof:* If the graph is not connected, then there are clearly an infinite number of solutions. Assume then that the graph is connected. In this case there exists a spanning tree. In the spirit of other proofs regarding the observability of multi-agent localization [31], [46], we proceed to show that the constraints along the spanning tree are sufficient for observability.

A minimum of the objective (25) corresponds to a maximum of $\sum_{(i,j) \in \mathcal{E}} \log \mathcal{W}_{\sigma_{ij}^2}(\boldsymbol{d}_{ij}^{-1} \boldsymbol{r}_i^{-1} \boldsymbol{r}_j)$; moreover, the maximum of a sum of functions cannot exceed the sum of the maxima, i.e.,

$$\max_{\{r_i\} \in \mathrm{SO}(2)^n} \sum_{(i,j) \in \mathcal{E}} \log \mathcal{W}_{\sigma_{ij}^2}(\boldsymbol{d}_{ij}^{-1} \boldsymbol{r}_i^{-1} \boldsymbol{r}_j) \leq \quad (26)$$
$$\sum_{(i,j) \in \mathcal{E}} \max_{\boldsymbol{r}_i, \boldsymbol{r}_j \in \mathrm{SO}(2)} \log \mathcal{W}_{\sigma_{ij}^2}(\boldsymbol{d}_{ij}^{-1} \boldsymbol{r}_i^{-1} \boldsymbol{r}_j) \doteq \hat{J}.$$

We will now show that, in the noiseless case, there exists a solution, say $\{\boldsymbol{r}_i^\star\} \in \mathrm{SO}(2)^n$, that attains the upper bound $\hat{J}$ (i.e., for $\{\boldsymbol{r}_i^\star\}$, eq. (26) holds with equality), and that such solution is unique. This implies that $\{\boldsymbol{r}_i^\star\} \in \mathrm{SO}(2)^n$ is the unique global maximum of the likelihood, and, in turn, it is the unique global minimum of the cost (25). For this purpose we notice that the peak of each wrapped Gaussian $\mathcal{W}_{\sigma_{ij}^2}(\boldsymbol{d}_{ij}^{-1} \boldsymbol{r}_i^{-1} \boldsymbol{r}_j)$ is attained at for $\boldsymbol{d}_{ij}^{-1} \boldsymbol{r}_i^{-1} \boldsymbol{r}_j = \mathbf{I}_2$ [45]. Therefore, we want to show that there exists a solution that satisfies

$$\boldsymbol{d}_{ij}^{-1} \boldsymbol{r}_i^{-1} \boldsymbol{r}_j = \mathbf{I}_2 \quad (27)$$

for all $(i,j) \in \mathcal{E}$ and that such solution is unique. Consider a spanning tree rooted at node 0, that has orientation $r_0 = \mathbf{I}_2$. This spanning tree exists for the hypothesis of connectivity. Consider a branch of the tree and call $i_1$ the first node encountered after the root, along this branch. Impose the condition (27) for the edge connecting node 0 and node $i_1$:

$$d_{0i_1}^{-1} r_0^{-1} r_{i_1} = \mathbf{I}_2.$$

Recalling that $r_0 = \mathbf{I}_2$, we conclude that the only orientation for node $i_1$ satisfying the previous equation is $r_{i_1}^\star = d_{0i_1}$ (i.e., the orientation is equal to the relative measurement w.r.t. the root). Call $i_2$ the second node along the spanning tree and impose the condition.

$$d_{i_1 i_2}^{-1} r_{i_1}^{-1} r_{i_2} = \mathbf{I}_2.$$

Again, this leads to a unique solution $r_{i_2}^\star = r_{i_1} d_{i_1 i_2} = d_{0i_1} d_{i_1 i_2}$. Repeating the same reasoning along all nodes in the branch of the spanning tree we obtain a unique orientation for each node that assure the satisfaction of equation (27) for all edges of the spanning tree. Now, we notice that, in a noiseless setup, the measurements corresponding to chords of the spanning tree are redundant, in the sense that if equation (27) is satisfied for all edges of the spanning tree, then, it needs be satisfied for the chords, too. Therefore, we found a (unique) solution $\{r_i^\star\} \in \mathrm{SO}(2)^n$ that is feasible for the original problem and attains $\hat{J}$, and this solution needs to be the global maximum of $\sum_{(i,j) \in \mathcal{E}} \log \mathcal{W}_{\sigma_{ij}^2}(d_{ij}^{-1} r_i^{-1} r_j)$, and, therefore, it is the unique global minimum of (25). ∎

Because of the observability condition of Proposition 3, in the rest of the paper we take the following assumption.

**Assumption 4.** *The graph $\mathcal{G}$ is connected*[1].

*A. Choosing coordinates*

As a first step, we make a choice of coordinates for the nodes orientation and measurements. We include this passage in the Problem Statement section since it leads to the problem formulation that is commonly adopted in literature.

We use the coordinates $\check{\boldsymbol{\theta}}^\circ \in (-\pi, +\pi]^n$ for the true unknown orientations, the coordinates $\check{\boldsymbol{\theta}} \in (-\pi, +\pi]^n$ for the optimization variables, and the coordinates $\check{\boldsymbol{\delta}} \in (-\pi, +\pi]^m$ for the relative measurements. Formally, these are defined as the principal logarithm of quantities that live on SO(2):

$$\check{\theta}_i^\circ \doteq \mathrm{Log}_0(r_i^\circ), \quad (28)$$
$$\check{\theta}_i \doteq \mathrm{Log}_0(r_i), \quad (29)$$
$$\check{\delta}_{ij} \doteq \mathrm{Log}_0(d_{ij}). \quad (30)$$

The $\check{\boldsymbol{\theta}}^\circ$ is the *estimand*, i.e., what we want to estimate (also later the mark "°" will label unknown quantities that need be estimated). We now want to write the objective function (25) as a least-squares cost, depending on the coordinates $\check{\boldsymbol{\theta}}$ and $\check{\boldsymbol{\delta}}$.

This is one of the first passages in which we must be careful. It is only possible to write the likelihood as a quadratic function if $\sigma_{ij}$ is small enough. As $\sigma_{ij}$ grows, the likelihood of the measurement $(i,j)$ tends to a constant that cannot be represented as a quadratic function.

---
[1]As in related work, the notion of connectivity is referred to the undirected version of $\mathcal{G}$; for this reason, it is also referred to as *weak* connectivity [3].

**Assumption 5.** *The uncertainty of a single relative orientation measurement does not "spill over" the $\pm \pi$ boundaries:*

$$3\sigma_{ij} \ll \pi. \quad (31)$$

This is not a strict assumption in robotics because angular measurements are usually much more precise than the given threshold.

In any case, if the uncertainty of some relative measurement is larger, we can use a simple trick. Because the convolution of two wrapped Gaussians is still a wrapped Gaussian (Lemma 2), we can replace one edge with a large variance $\sigma_{ij}^2$ with two edges with smaller variances whose sum is $\sigma_{ij}^2$. Consequently, we can assume without loss of generality that Assumption 5 is satisfied.

Assumption 5 allows to write the log-likelihood function as a quadratic function.

**Lemma 6** (Quadratic approximation of wrapped Gaussian likelihood). *If Assumption 5 is satisfied, then*

$$-\log \mathcal{W}_{\sigma_{ij}^2}(\boldsymbol{\varepsilon}_{ij}) \approx \frac{|\mathrm{Log}_0(\boldsymbol{\varepsilon}_{ij})|^2}{2\sigma_{ij}^2} + \log\left(\sigma_{ij}\sqrt{2\pi}\right). \quad (32)$$

*Proof:* The proof is a direct consequence of the expression of the wrapped Gaussian distribution (19):

$$\mathcal{W}_{\sigma_{ij}^2}(\boldsymbol{\varepsilon}_{ij}) = \frac{1}{\sigma_{ij}\sqrt{2\pi}} \sum_{k=-\infty}^{+\infty} \exp\left(\frac{-(\mathrm{Log}_0(\boldsymbol{\varepsilon}_{ij}) + 2\pi k)^2}{2\sigma_{ij}^2}\right)$$
$$\overset{3\sigma_{ij} \ll \pi}{\approx} \frac{1}{\sigma_{ij}\sqrt{2\pi}} \exp\left(\frac{-(\mathrm{Log}_0(\boldsymbol{\varepsilon}_{ij}))^2}{2\sigma_{ij}^2}\right),$$

which implies (32). ∎

Lemma 6 allows writing the likelihood using only the coordinates $\check{\boldsymbol{\theta}}, \check{\boldsymbol{\delta}}$ and the modulus operation $\langle \cdot \rangle_{2\pi}$.

**Lemma 7** (Quadratic approximation in coordinates). *If Assumption 5 is satisfied, then*

$$-\log \mathcal{W}_{\sigma_{ij}^2}(d_{ij}^{-1} r_i^{-1} r_j) \simeq \frac{1}{2\sigma_{ij}^2}\left|\langle \check{\theta}_j - \check{\theta}_i - \check{\delta}_{ij} \rangle_{2\pi}\right|^2 + c, \quad (33)$$

*with the constant $c$ equal to $\log\left(\sigma_{ij}\sqrt{2\pi}\right)$.*

*Proof:* From (32), and recalling the definition of $d_{ij}$ in (21) it follows that

$$-\log \mathcal{W}_{\sigma_{ij}^2}(d_{ij}^{-1} r_i^{-1} r_j) \simeq \frac{1}{2\sigma_{ij}^2}|\mathrm{Log}_0(d_{ij}^{-1} r_i^{-1} r_j)|^2 + c. \quad (34)$$

The rest of the proof consists of algebraic manipulation based

on properties that we have already introduced:

$$|\text{Log}_0(\boldsymbol{d}_{ij}^{-1}\boldsymbol{r}_i^{-1}\boldsymbol{r}_j)|^2$$
$$= \left|\langle \text{Log}(\boldsymbol{d}_{ij}^{-1}\boldsymbol{r}_i^{-1}\boldsymbol{r}_j)\rangle_{2\pi}\right|^2$$
(Using the property of $\text{Log}_0$ in (17).)
$$= \left|\langle \text{Log}(\boldsymbol{d}_{ij}^{-1}) + \text{Log}(\boldsymbol{r}_i^{-1}) + \text{Log}(\boldsymbol{r}_j)\rangle_{2\pi}\right|^2$$
(Using the property of SO(2) in (15).)
$$= \left|\langle -\text{Log}(\boldsymbol{d}_{ij}) - \text{Log}(\boldsymbol{r}_i) + \text{Log}(\boldsymbol{r}_j)\rangle_{2\pi}\right|^2$$
(Using the property of SO(2) in (14).)
$$= \left|\langle -\langle\text{Log}(\boldsymbol{d}_{ij})\rangle_{2\pi} - \langle\text{Log}(\boldsymbol{r}_i)\rangle_{2\pi} + \langle\text{Log}(\boldsymbol{r}_j)\rangle_{2\pi}\rangle_{2\pi}\right|^2$$
(Using the property (11).)
$$= \left|\langle \text{Log}_0(\boldsymbol{d}_{ij}) - \text{Log}_0(\boldsymbol{r}_i) + \text{Log}_0(\boldsymbol{r}_j)\rangle_{2\pi}\right|^2$$
(Using the property (17).)
$$= \left|\langle \check{\delta}_{ij} + \check{\theta}_j - \check{\theta}_i\rangle_{2\pi}\right|^2.$$
(Using the definition (28).) ■

Based on Lemma 7, we can restate Problem 2 as the minimization of a quadratic function (but still with the $\langle\cdot\rangle_{2\pi}$ nonlinearity).

**Problem 3** (Angular coordinates formulation of maximum-likelihood orientation estimation). *Given the observations $\check{\delta}_{ij} \in (-\pi, +\pi]$, for $(i,j) \in \mathcal{E}$, and the corresponding variances $\sigma_{ij} > 0$, find the set of minimizers $S^3 \subset (-\pi, +\pi]^n$ that satisfies*

$$S^3 = \underset{\check{\boldsymbol{\theta}} \in (-\pi,+\pi]^n}{\arg\min} \sum_{(i,j)\in\mathcal{E}} \frac{1}{\sigma_{ij}^2} \left|\left\langle \check{\theta}_j - \check{\theta}_i - \check{\delta}_{ij}\right\rangle_{2\pi}\right|^2, \quad (35)$$

*having fixed the first node orientation to $\check{\theta}_0 = 0$.*

For clarity, and without loss of generality, the constant term in (33) is omitted in (35).

This formulation is the same as the one used in related work [15], [38], however, rather than tacitly assuming that the uncertainty of each measurement does not "spill over" the $\pi$ boundary, we will be explicit on how to deal with the modulus operation.

*1) From Problem 3 to Problem 2:* The conversion between the solutions of the two problems is just a change of coordinates:

$$S^2 = \text{Exp}(S^3), \quad (36)$$
$$S^3 = \text{Log}_0(S^2). \quad (37)$$

*2) Symmetries of Problem 3:* If Assumption 5 is satisfied, Problem 3 is just a restatement in different coordinates of Problem 2, so they have the same symmetries. In the noiseless case, the solution is unique (Proposition 3). We still do not know what happens in the noisy case, but we can anticipate (Proposition 14) that, for general data, the solution is unique.

### B. Why is Problem 3 hard?

Besides the exponential mapping, Problem 3 appears to be close to a standard linear estimation problem. Actually, this is not the case. In standard linear estimation the cost function to minimize (to obtain the maximum likelihood estimate) is quadratic, hence it has a single minimum. Convexity, in the linear case, guarantees that (most) iterative optimization algorithms can efficiently compute the global minimum. The cost function (35), instead, is non-convex and has several local minima (Figure 2).

Moreover, we will show in Section V that the problem is actually even more challenging: even if one is able to compute a global minimum of the cost function (35) (which quantifies the *data error*, i.e., the mismatch between the expected and the actual measurements) this does not guarantee that the corresponding orientation estimate is accurate (in the sense of having small *estimation error*, which is the mismatch between the estimate and the actual orientations).

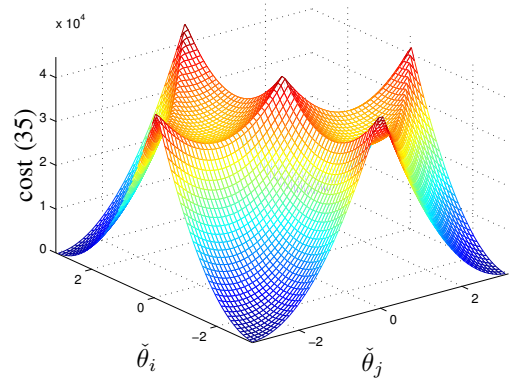

Fig. 2. Value of the cost (35) as a function of two robot orientations, while the remaining orientations are kept fixed. The data is taken from the INTEL dataset (Section VIII-A).

## IV. MAXIMUM LIKELIHOOD ESTIMATION ON SO(2): FROM ANGLES TO INTEGERS

This section shows that the nonlinear, nonconvex, constrained Problem 3 is equivalent to an unconstrained quadratic integer optimization problem.

Table II shows our strategy: we will convert Problem 3, which is defined on $(-\pi, +\pi]^n$, where $n$ is the number of nodes, through a series of intermediate formulations, until we arrive at Problem 7, which is defined on the integers $\mathbb{Z}^\ell$, where $\ell$ is the number of *cycles* in the graph. The final result of this section, Theorem 16 on page 13, says that the solution of Problem 3 is almost surely unique, and that we can obtain such solution by solving Problem 7.

The road towards that result is quite long, and it goes through different reformulations of the optimization problem, defined on different domains, according to two principles: sometimes it is convenient to solve a simpler problem in a larger space; and sometimes it is possible to shrink the domain.

The set $S^i$ is the set of minimizers of the $i$-th problem. At each step on the road we characterize the properties of this set, and in particular we derive its symmetry group. This is important because some of the intermediate formulations, namely Problem 4 to Problem 6, have multiple solutions.

These are not ambiguities of the original problem, but rather artifacts of our choice of using a redundant representation. Describing the symmetry groups formally makes us able to quantify the redundancy and make sure we are not forgetting any solution or introducing new spurious solutions.

The maps $\varphi_j^i$ project the solutions $S^j$ to the solutions $S^i$. These maps are not all invertible. The final conclusion, however, is that the composition of these maps is surjective, so that we are sure that, by solving Problem 7 and then applying the composition, we recover all solutions of Problem 3.

*Structure of the section*: Section IV-A formulates Problem 4, whose optimization variable $\boldsymbol{\theta}$ is defined on the set of real numbers (hence not constrained in $(-\pi, +\pi]^n$, as it is for Problem 3). Here, we work on a larger domain, and hence we introduce an additional ambiguity, which is that each entry of $\boldsymbol{\theta}$ is determined only modulo $2\pi$.

Section IV-B formulates Problem 5, which is defined on $(\boldsymbol{\theta}, \boldsymbol{k}) \in \mathbb{R}^n \times \mathbb{Z}^m$, $m$ being the number of edges in the graph. The new variable $\boldsymbol{k} \in \mathbb{Z}^m$ can be interpreted as compensation terms that we introduce to keep track of "angular excess".

Section IV-C shows that, given the value of $\boldsymbol{k}$, the value of $\boldsymbol{\theta}$ can be recovered in a closed form using linear estimation: in fact, if we knew $\boldsymbol{k}$, the problem would be linear.

Section IV-D formulates Problem 6, which is defined only on $\boldsymbol{k} \in \mathbb{Z}^m$. The insight is that the integer and the real part of the problem can be solved separately, thus allowing a two-stage optimization procedure.

Section IV-E formulates Problem 7, which is defined on an integer variable $\boldsymbol{\gamma} \in \mathbb{Z}^\ell$. While $\boldsymbol{k}$ lives on the edges, $\boldsymbol{\gamma}$ lives on the cycles of the graph. We show that $\boldsymbol{\gamma}$ is a minimal parametrization for our problem.

Finally, Section IV-G puts together the chain of implications, and shows that the solution set $S^7$ can be mapped surjectively to $S^3$, and we can easily compute $S^3$ once we know $S^7$.

### A. Real-valued formulation

The first step is the reformulation of Problem 3 as an *unconstrained* optimization problem for real variables $\boldsymbol{\theta} \in \mathbb{R}^n$.

**Problem 4** (Real-valued formulation of maximum-likelihood orientation estimation). *Given the observations $\check{\delta}_{ij} \in (-\pi, +\pi]$, $(i,j) \in \mathcal{E}$ and the corresponding variances $\sigma_{ij} > 0$, find the set $S^4 \subset \mathbb{R}^n$ that satisfies*

$$S^4 = \underset{\boldsymbol{\theta} \in \mathbb{R}^n}{\arg\min} \sum_{(i,j) \in \mathcal{E}} \frac{1}{\sigma_{ij}^2} |\langle \theta_j - \theta_i - \check{\delta}_{ij} \rangle_{2\pi}|^2, \quad (38)$$

*having fixed $\theta_0 = 0$.*

We use a check mark ( $\check{\;}$ ) to label quantities belonging to $(-\pi, +\pi]$ in order to distinguish them from quantities in $\mathbb{R}^n$. For example, $\check{\boldsymbol{\theta}} \in (-\pi, +\pi]$ and $\boldsymbol{\theta} \in \mathbb{R}^n$.

*1) From Problem 4 to Problem 3:* While Problem 4 has multiple solutions due to the symmetry, they are all equivalent when they are projected down to the manifold using the exponential map

$$\varphi_4^3 : \mathbb{R}^n \rightarrow (-\pi, +\pi]^n \quad (39)$$
$$\boldsymbol{\theta} \mapsto \langle \boldsymbol{\theta} \rangle_{2\pi}. \quad (40)$$

More formally, the set $S^3$ can be obtained from $S^4$ by applying the map $\varphi_4^3$, as stated in the following proposition.

**Proposition 8.** $S^3 = \varphi_4^3(S^4)$.

*Proof:* In order to prove the claim, we have to show that (i) for any $\boldsymbol{\theta}^\star \in S^4$, the variable $\langle \boldsymbol{\theta}^\star \rangle_{2\pi}$ is in $S^3$, and (ii) for any solution $\check{\boldsymbol{\theta}}^\star \in S^3$ there exists at least one $\boldsymbol{\theta}^\star \in S^4$, such that $\check{\boldsymbol{\theta}}^\star = \langle \boldsymbol{\theta}^\star \rangle_{2\pi}$. Let us start from the first implication. We note that problems (38) and (35) have the same objective and we use the notation $J(\boldsymbol{x})$, to denote the value of this objective for a given vector $\boldsymbol{x}$; therefore, for any $\boldsymbol{\theta}^\star \in S^4$, $J(\boldsymbol{\theta}^\star)$ is the optimal objective of (38), while for any $\check{\boldsymbol{\theta}}^\star \in S^3$, $J(\check{\boldsymbol{\theta}}^\star)$ is the optimal objective of (35). Therefore, we want to prove that for any $\boldsymbol{\theta}^\star \in S^4$, it holds $J(\langle \boldsymbol{\theta}^\star \rangle_{2\pi}) = J(\check{\boldsymbol{\theta}}^\star)$, i.e., $\langle \boldsymbol{\theta}^\star \rangle_{2\pi}$ attains the optimal objective in (35). Let us prove this equality. First of all we notice that $\langle \boldsymbol{\theta}^\star \rangle_{2\pi}$ satisfies the constraints in (35), by definition of exponential map; therefore, $\langle \boldsymbol{\theta}^\star \rangle_{2\pi}$ is a feasible solution for problem (35) and it must hold (a) $J(\langle \boldsymbol{\theta}^\star \rangle_{2\pi}) \geq J(\check{\boldsymbol{\theta}}^\star)$ ($\langle \boldsymbol{\theta}^\star \rangle_{2\pi}$ cannot be better than the optimal solution). Now we notice that problem (38) is the same as (35) but without the constraints for the angles to belong to $(-\pi, +\pi]$; therefore, it must hold (b) $J(\boldsymbol{\theta}^\star) \leq J(\check{\boldsymbol{\theta}}^\star)$ (relaxing constraints can only improve the objective). Finally, using property (11), we can easily see that (c) $J(\langle \boldsymbol{\theta}^\star \rangle_{2\pi}) = J(\boldsymbol{\theta}^\star)$. Combining relations (a), (b), and (c) we obtain:

$$J(\check{\boldsymbol{\theta}}^\star) \overset{(a)}{\leq} J(\langle \boldsymbol{\theta}^\star \rangle_{2\pi}) \overset{(c)}{=} J(\boldsymbol{\theta}^\star) \overset{(b)}{\leq} J(\check{\boldsymbol{\theta}}^\star) \quad (41)$$

which implies $J(\langle \boldsymbol{\theta}^\star \rangle_{2\pi}) = J(\check{\boldsymbol{\theta}}^\star)$, proving the first statement. For the second claim, we notice that relation (41) also implies $J(\boldsymbol{\theta}^\star) = J(\check{\boldsymbol{\theta}}^\star)$. This fact tells us that any solution $\check{\boldsymbol{\theta}}^\star \in S^3$ also attains the minimum of (38), i.e., $\check{\boldsymbol{\theta}}^\star \in S^4$. Moreover, since $\check{\boldsymbol{\theta}}^\star \in (-\pi, +\pi]^n$, it holds that $\check{\boldsymbol{\theta}}^\star = \langle \check{\boldsymbol{\theta}}^\star \rangle_{2\pi}$, therefore, for each $\check{\boldsymbol{\theta}}^\star \in S^3$, we can choose an element of $S^4$, namely $\boldsymbol{\theta}^\star = \check{\boldsymbol{\theta}}^\star$, such that $\langle \boldsymbol{\theta}^\star \rangle_{2\pi} = \check{\boldsymbol{\theta}}^\star$ belongs to $S^3$, which proves the second claim. ∎

Using this result, we can solve the unconstrained Problem 4, obtaining a solution $\boldsymbol{\theta}^\star \in \mathbb{R}^n$, and then compute an optimal solution of Problem 3 as $\check{\boldsymbol{\theta}}^\star \doteq \langle \boldsymbol{\theta}^\star \rangle_{2\pi} \in (-\pi, +\pi]^n$.

*2) Symmetries of Problem 4:* Note that this problem has more solutions than Problem 3. This is an artifact of the real-valued parametrization. In particular, if $\boldsymbol{\theta} \in \mathbb{R}^n$ is a solution, also $\boldsymbol{\theta} - 2\pi \boldsymbol{p}$ is a solution, for any integer vector $\boldsymbol{p} \in \mathbb{Z}^n$. Note, instead, that if $\boldsymbol{\theta}$ is a solution, not necessarily $\boldsymbol{\theta} + s\mathbf{1}_n$ is a solution, because we have fixed the first node (unless $s$ is a multiple of $2\pi$).

### B. Mixed-integer formulation

Problem 4 is an optimization problem in real variables, but its residual errors are still nonlinear and difficult to minimize. We now get to the core idea of this paper: instead of solving a *nonlinear* problem in *real* variables, we choose to solve a *linear* problem in *mixed (integer and real)* variables.

The "trick" is that one can get rid of the exponential map in the expression of the residuals

$$|\langle \theta_j - \theta_i - \check{\delta}_{ij} \rangle_{2\pi}|^2, \quad (42)$$

by the introduction of a $2\pi$ factor that depends on an integer $k_{ij}$:
$$|\theta_j - \theta_i - \check{\delta}_{ij} + 2\pi k_{ij}|^2. \tag{43}$$

More precisely, by using the property (10), it holds that
$$|\langle \theta_j - \theta_i - \check{\delta}_{ij} \rangle_{2\pi}|^2 = \min_{k_{ij}} |\theta_j - \theta_i - \check{\delta}_{ij} + 2\pi k_{ij}|^2. \tag{44}$$

With the introduction of the regularization terms, the error function (38) can be written as
$$\sum_{(i,j) \in \mathcal{E}} \tfrac{1}{\sigma_{ij}^2} |\theta_j - \theta_i - \check{\delta}_{ij} + 2\pi k_{ij}|^2. \tag{45}$$

This can be written in a more compact form using the *reduced incidence matrix* $A$ of the graph (Section II-A). Suppose that there is an ordering of the edges from 1 to $m$, so that the measurements can be written as an $m$-dimensional vector
$$\check{\boldsymbol{\delta}} \in (-\pi, +\pi]^m. \tag{46}$$

Define accordingly the *regularization vector*
$$\boldsymbol{k} = (k_1 \ k_2 \ \cdots \ k_m)^\mathsf{T} \in \mathbb{Z}^m, \tag{47}$$

and the *measurement covariance*
$$\boldsymbol{P}_\delta = \mathrm{diag}(\sigma_1^2, \sigma_2^2, \ldots, \sigma_m^2) \in \mathbb{R}^{m \times m}. \tag{48}$$

If $\overline{\boldsymbol{A}} \in \{-1, 0, +1\}^{(n+1) \times m}$ is the incidence matrix of the graph, its transpose $\overline{\boldsymbol{A}}^\mathsf{T} \in \{-1, 0, +1\}^{m \times (n+1)}$ is a linear operator that transforms a quantity on the nodes (in $\mathbb{R}^n$) to a quantity on the edges. Using this vector notation, and recalling the definition of the reduced incidence matrix given in Section II-A, we obtain the following reformulation that uses both integer and real-valued variables.

**Problem 5** (Mixed-integer formulation of maximum-likelihood orientation estimation). *Given the vector $\check{\boldsymbol{\delta}} \in (-\pi, +\pi]^m$ and the diagonal positive definite matrix $\boldsymbol{P}_\delta \in \mathbb{R}^{m \times m}$, find the set of minimizers $S^5 \subset \mathbb{R}^n \times \mathbb{Z}^m$ that satisfy*
$$S^5 = \underset{(\boldsymbol{\theta},\boldsymbol{k}) \in \mathbb{R}^n \times \mathbb{Z}^m}{\arg\min} \left\| \boldsymbol{A}^\mathsf{T} \boldsymbol{\theta} - \check{\boldsymbol{\delta}} + 2\pi \boldsymbol{k} \right\|_{\boldsymbol{P}_\delta^{-1}}^2. \tag{49}$$

Problem 5 has quadratic objective and includes both continuous and discrete variables, hence belongs to the class of mixed-integer convex programs [36].

*1) From Problem 5 to Problem 4:* From the solutions of Problem 5 we can obtain the solutions to Problem 4 using the projection map
$$\varphi_5^4 : \mathbb{R}^n \times \mathbb{Z}^m \to \mathbb{R}^n \tag{50}$$
$$(\boldsymbol{\theta}, \boldsymbol{k}) \mapsto \boldsymbol{\theta}. \tag{51}$$

**Proposition 9.** $\varphi_5^4(S^5) = S^4$.

*Proof:* We show how to transform Problem 4 into Problem 5. Using property (10) we can rewrite (38) as
$$\min_{\boldsymbol{\theta} \in \mathbb{R}^n} \sum_{(i,j) \in \mathcal{E}} \tfrac{1}{\sigma_{ij}^2} \min_{k_{ij} \in \mathbb{Z}} |\theta_j - \theta_i - \check{\delta}_{ij} + 2\pi k_{ij}|^2,$$

which corresponds to
$$\min_{\boldsymbol{\theta} \in \mathbb{R}^n, \ k_{ij} \in \mathbb{Z}, (i,j) \in \mathcal{E}} \sum_{(i,j) \in \mathcal{E}} \tfrac{1}{\sigma_{ij}^2} |\theta_j - \theta_i - \check{\delta}_{ij} + 2\pi k_{ij}|^2. \tag{52}$$

Therefore, $k_{ij}$ are only slack variables that substitute the exponential map, and finding the optimal $\boldsymbol{\theta}$ for (52) is the same as finding the optimal $\boldsymbol{\theta}$ for Problem 4. Then, we conclude the proof by noting that in Problem 5, we only wrote (52) using a compact matrix notation. ∎

*2) Symmetries of Problem 5:* Note that we are now working on a larger space $\mathbb{R}^n \times \mathbb{Z}^m$: we are overparametrizing the problem in order to make the corresponding cost function quadratic. Therefore, we might have enlarged the number of solutions. In fact, we introduced the following symmetry. For any vector $\boldsymbol{p} \in \mathbb{R}^n$, such that $\boldsymbol{A}^\mathsf{T} \boldsymbol{p}$ is an integer vector, the following transformation leaves the error function invariant:
$$(\boldsymbol{\theta}, \boldsymbol{k}) \mapsto (\boldsymbol{\theta} - 2\pi \boldsymbol{p}, \boldsymbol{k} + \boldsymbol{A}^\mathsf{T} \boldsymbol{p}). \tag{53}$$

Because $\boldsymbol{A}^\mathsf{T}$ has full column rank, this is the only symmetry. We note that, for the particular structure of the matrix $\boldsymbol{A}^\mathsf{T}$ (having $-1$ and $+1$ as only nonzero elements in each row, and having some rows with a single nonzero element being either $-1$ or $+1$) only integer vectors $\boldsymbol{p} \in \mathbb{Z}^n$ may produce an integer $\boldsymbol{A}^\mathsf{T} \boldsymbol{p}$.

### C. Solving for $\boldsymbol{\theta}$ given known $\boldsymbol{k}$

Before further manipulation of Problem 5, we introduce formulas for the estimation of $\boldsymbol{\theta}$ in the case $\boldsymbol{k}$ was known.

Once the regularization vector $\boldsymbol{k}$ is known, say $\boldsymbol{k} = \bar{\boldsymbol{k}}$, the optimization problem becomes an unconstrained quadratic problem in $\boldsymbol{\theta} \in \mathbb{R}^n$:
$$\min_{\boldsymbol{\theta} \in \mathbb{R}^n} \left\| \boldsymbol{A}^\mathsf{T} \boldsymbol{\theta} - \check{\boldsymbol{\delta}} + 2\pi \bar{\boldsymbol{k}} \right\|_{\boldsymbol{P}_\delta^{-1}}^2. \tag{54}$$

This problem can be solved in a closed form. Denote by $\boldsymbol{\theta}^{\star|\boldsymbol{k}}$ the optimal $\boldsymbol{\theta}$ for a fixed $\boldsymbol{k}$:
$$\boldsymbol{\theta}^{\star|\boldsymbol{k}} = (\boldsymbol{A} \boldsymbol{P}_\delta^{-1} \boldsymbol{A}^\mathsf{T})^{-1} \boldsymbol{A} \boldsymbol{P}_\delta^{-1} (\check{\boldsymbol{\delta}} - 2\pi \boldsymbol{k}). \tag{55}$$

### D. Separating the integer-valued and the real-valued problems

This section shows that the cost function (49) is separable into two terms, enabling a two-stage optimization in which the cost is first optimized with respect to $\boldsymbol{k}$ and then the optimal choice of $\boldsymbol{\theta}$ is computed in closed-form.

The following lemma gives the separability result. It uses a cycle basis matrix $\boldsymbol{C}$, and it is valid for *any* cycle basis. In Section VI-C we discuss the implications of the choice of a particular cycle basis matrix.

**Lemma 10.** *For any given cycle basis matrix $\boldsymbol{C}$, minimizing the cost* (49) *is the same as minimizing*
$$\left\| \boldsymbol{\theta} - \boldsymbol{\theta}^{\star|\boldsymbol{k}} \right\|_{(\boldsymbol{A} \boldsymbol{P}_\delta^{-1} \boldsymbol{A}^\mathsf{T})}^2 + \left\| 2\pi \boldsymbol{C} \boldsymbol{k} - \boldsymbol{C} \check{\boldsymbol{\delta}} \right\|_{(\boldsymbol{C} \boldsymbol{P}_\delta \boldsymbol{C}^\mathsf{T})^{-1}}^2, \tag{56}$$

*where $\boldsymbol{\theta}^{\star|\boldsymbol{k}}$ is a function of $\boldsymbol{k}$ and is given in* (55).

*Proof:* The proof consists of straightforward algebraic manipulations. For compactness, we name the matrices
$$\boldsymbol{X} \doteq \boldsymbol{A}^\mathsf{T} (\boldsymbol{A} \boldsymbol{P}_\delta^{-1} \boldsymbol{A}^\mathsf{T})^{-1} \boldsymbol{A}, \quad \boldsymbol{Y} \doteq \boldsymbol{C}^\mathsf{T} (\boldsymbol{C} \boldsymbol{P}_\delta \boldsymbol{C}^\mathsf{T})^{-1} \boldsymbol{C}. \tag{57}$$

Expand the term $\| \boldsymbol{A}^\mathsf{T} \boldsymbol{\theta} - \check{\boldsymbol{\delta}} + 2\pi \boldsymbol{k} \|_{\boldsymbol{P}_\delta^{-1}}^2$ in (49) to obtain
$$\begin{aligned} &\| \boldsymbol{\theta} \|_{\boldsymbol{A} \boldsymbol{P}_\delta^{-1} \boldsymbol{A}^\mathsf{T}}^2 - 2 \check{\boldsymbol{\delta}}^\mathsf{T} \boldsymbol{P}_\delta^{-1} \boldsymbol{A}^\mathsf{T} \boldsymbol{\theta} + 4\pi \boldsymbol{k}^\mathsf{T} \boldsymbol{P}_\delta^{-1} \boldsymbol{A}^\mathsf{T} \boldsymbol{\theta} \\ &+ 4\pi^2 \| \boldsymbol{k} \|_{\boldsymbol{P}_\delta^{-1}}^2 - 4\pi \check{\boldsymbol{\delta}}^\mathsf{T} \boldsymbol{P}_\delta^{-1} \boldsymbol{k} + \| \check{\boldsymbol{\delta}} \|_{\boldsymbol{P}_\delta^{-1}}^2. \end{aligned} \tag{58}$$

Because $\|\check{\boldsymbol{\delta}}\|^2_{P_\delta^{-1}}$ does not depend on the optimization variables, minimizing (58) is the same as minimizing

$$f_1(\boldsymbol{\theta}, \boldsymbol{k}) \doteq \|\boldsymbol{\theta}\|^2_{AP_\delta^{-1}A^\mathsf{T}} - 2\check{\boldsymbol{\delta}}^\mathsf{T} P_\delta^{-1} A^\mathsf{T} \boldsymbol{\theta} + 4\pi \boldsymbol{k}^\mathsf{T} P_\delta^{-1} A^\mathsf{T} \boldsymbol{\theta}$$
$$+ 4\pi^2 \|\boldsymbol{k}\|^2_{P_\delta^{-1}} - 4\pi \check{\boldsymbol{\delta}}^\mathsf{T} P_\delta^{-1} \boldsymbol{k}. \quad (59)$$

To show that minimizing $f_1(\boldsymbol{\theta}, \boldsymbol{k})$ in (59) is the same as minimizing (56), we rewrite the latter as

$$\|\boldsymbol{\theta}\|^2_{AP_\delta^{-1}A^\mathsf{T}} - 2(\boldsymbol{\theta}^{\star|\boldsymbol{k}})^\mathsf{T}(AP_\delta^{-1}A^\mathsf{T})\boldsymbol{\theta}$$
$$+ \|\boldsymbol{\theta}^{\star|\boldsymbol{k}}\|^2_{AP_\delta^{-1}A^\mathsf{T}} + 4\pi^2 \|\boldsymbol{k}\|^2_Y - 4\pi \check{\boldsymbol{\delta}}^\mathsf{T} Y \boldsymbol{k} + \|\check{\boldsymbol{\delta}}\|^2_Y \quad (60)$$

Using the matrices $X$ and $Y$, and recalling the definition of $\boldsymbol{\theta}^{\star|\boldsymbol{k}}$ in (55), (60) can be rewritten as

$$\|\boldsymbol{\theta}\|^2_{AP_\delta^{-1}A^\mathsf{T}} - 2(\check{\boldsymbol{\delta}} - 2\pi\boldsymbol{k})^\mathsf{T} P_\delta^{-1} A^\mathsf{T} \boldsymbol{\theta}$$
$$+ \|\check{\boldsymbol{\delta}} - 2\pi\boldsymbol{k}\|^2_{P_\delta^{-1}XP_\delta^{-1}} \quad (61)$$
$$+ 4\pi^2 \|\boldsymbol{k}\|^2_Y - 4\pi \check{\boldsymbol{\delta}}^\mathsf{T} Y \boldsymbol{k} + \|\check{\boldsymbol{\delta}}\|^2_Y.$$

We can further develop the previous expression, computing the products in the second and in the third summand:

$$\|\boldsymbol{\theta}\|^2_{AP_\delta^{-1}A^\mathsf{T}} - 2\check{\boldsymbol{\delta}}^\mathsf{T} P_\delta^{-1} A^\mathsf{T} \boldsymbol{\theta}$$
$$+ 4\pi \boldsymbol{k}^\mathsf{T} P_\delta^{-1} A^\mathsf{T} \boldsymbol{\theta} + \|\check{\boldsymbol{\delta}}\|^2_{P_\delta^{-1}XP_\delta^{-1}}$$
$$+ 4\pi^2 \|\boldsymbol{k}\|^2_{P_\delta^{-1}XP_\delta^{-1}} - 4\pi \check{\boldsymbol{\delta}}^\mathsf{T} P_\delta^{-1} X P_\delta^{-1} \boldsymbol{k} \quad (62)$$
$$+ 4\pi^2 \|\boldsymbol{k}\|^2_Y - 4\pi \check{\boldsymbol{\delta}}^\mathsf{T} Y \boldsymbol{k} + \|\check{\boldsymbol{\delta}}\|^2_Y.$$

Because $\|\check{\boldsymbol{\delta}}\|^2_{P_\delta^{-1}XP_\delta^{-1}}$ and $\|\check{\boldsymbol{\delta}}\|^2_Y$ are constant and do not depend on $\boldsymbol{\theta}$ and $\boldsymbol{k}$, minimizing (62) is the same as minimizing

$$f_2(\boldsymbol{\theta}, k) \doteq \|\boldsymbol{\theta}\|^2_{AP_\delta^{-1}A^\mathsf{T}} - 2\check{\boldsymbol{\delta}}^\mathsf{T} P_\delta^{-1} A^\mathsf{T} \boldsymbol{\theta}$$
$$+ 4\pi \boldsymbol{k}^\mathsf{T} P_\delta^{-1} A^\mathsf{T} \boldsymbol{\theta} + 4\pi^2 \|\boldsymbol{k}\|^2_{P_\delta^{-1}XP_\delta^{-1}}$$
$$- 4\pi \check{\boldsymbol{\delta}}^\mathsf{T} P_\delta^{-1} X P_\delta^{-1} \boldsymbol{k} + 4\pi^2 \|\boldsymbol{k}\|^2_Y - 4\pi \check{\boldsymbol{\delta}}^\mathsf{T} Y \boldsymbol{k}. \quad (63)$$

Comparing (59) and (63), one concludes that the first three terms in $f_1(\boldsymbol{\theta}, \boldsymbol{k})$ and $f_2(\boldsymbol{\theta}, \boldsymbol{k})$ coincide, and it only remains to show equality for the last terms. Rewrite (63) as

$$\|\boldsymbol{\theta}\|^2_{AP_\delta^{-1}A^\mathsf{T}} - 2\check{\boldsymbol{\delta}}^\mathsf{T} P_\delta^{-1} A^\mathsf{T} \boldsymbol{\theta} + 4\pi \boldsymbol{k}^\mathsf{T} P_\delta^{-1} A^\mathsf{T} \boldsymbol{\theta}$$
$$+ 4\pi^2 \boldsymbol{k}^\mathsf{T}(P_\delta^{-1} X P_\delta^{-1} + Y)\boldsymbol{k} \quad (64)$$
$$- 4\pi \check{\boldsymbol{\delta}}^\mathsf{T}(P_\delta^{-1} X P_\delta^{-1} + Y)\boldsymbol{k}.$$

Since $P_\delta$ is positive definite, the technical result of Lemma 23 (in appendix) implies that

$$P_\delta^{-1} = P_\delta^{-1} A^\mathsf{T}(AP_\delta^{-1}A^\mathsf{T})^{-1} AP_\delta^{-1} + C^\mathsf{T}(CP_\delta C^\mathsf{T})^{-1} C. \quad (65)$$

Hence $P_\delta^{-1} = P_\delta^{-1} X P_\delta^{-1} + Y$, and (64) becomes

$$\|\boldsymbol{\theta}\|^2_{AP_\delta^{-1}A^\mathsf{T}} - 2\check{\boldsymbol{\delta}}^\mathsf{T} P_\delta^{-1} A^\mathsf{T} \boldsymbol{\theta} + 4\pi \boldsymbol{k}^\mathsf{T} P_\delta^{-1} A^\mathsf{T} \boldsymbol{\theta} \quad (66)$$
$$+ 4\pi^2 \boldsymbol{k}^\mathsf{T} P_\delta^{-1} \boldsymbol{k} - 4\pi \check{\boldsymbol{\delta}}^\mathsf{T} P_\delta^{-1} \boldsymbol{k},$$

which can be easily seen to coincide with (59). Since the objective functions $f_1(\boldsymbol{\theta}, k)$ and $f_2(\boldsymbol{\theta}, k)$ coincide, problems (49) and (56) have the same solutions. ∎

A consequence of writing the error function as in (56) is that a separability principle holds: we can obtain the maximum-likelihood solution using a two-stage approach: first we estimate the $\boldsymbol{k}$, and then we estimate $\boldsymbol{\theta}$ given $\boldsymbol{k}$. This aspect is formalized later, in Proposition 11. Intuitively, the cost function (56) comprises two terms, the first that can be made equal to zero choosing $\boldsymbol{\theta} = \boldsymbol{\theta}^{\star|\boldsymbol{k}}$, and the second, that does not depend on $\boldsymbol{\theta}$ and can be minimized by working on $\boldsymbol{k}$. Since we already have a closed-form expression for $\boldsymbol{\theta}$ given $\boldsymbol{k}$ the only problem that we have to solve is finding $\boldsymbol{k}$.

**Problem 6** (Integer formulation of maximum-likelihood orientation estimation in edge space). *Given the vector $\check{\boldsymbol{\delta}} \in (-\pi, +\pi]^m$ and the diagonal positive definite matrix $P_\delta \in \mathbb{R}^{m \times m}$, and a cycle basis matrix $C \in \mathbb{Z}^{\ell \times m}$, find the set of minimizers $S^6 \subset \mathbb{Z}^m$ that satisfy*

$$S^6 = \underset{\boldsymbol{k} \in \mathbb{Z}^m}{\arg\min} \left\| C\boldsymbol{k} - \tfrac{1}{2\pi} C\check{\boldsymbol{\delta}} \right\|^2_{(CP_\delta C^\mathsf{T})^{-1}}. \quad (67)$$

Notice that (67) is the same as the second summand in (56), and we only rearranged the $2\pi$ term.

*1) From Problem 6 to Problem 5:* Proposition 11 assures that solving Problem 6 is the same as solving Problem 5, if we convert the solutions using the map

$$\varphi^5_6 : \mathbb{Z}^m \to \mathbb{R}^n \times \mathbb{Z}^m \quad (68)$$
$$\boldsymbol{k} \mapsto (\boldsymbol{\theta}^{\star|\boldsymbol{k}}, \boldsymbol{k}). \quad (69)$$

**Proposition 11.** $\varphi^5_6(S^6) = S^5$.

*Proof:* Since $AP_\delta^{-1}A^\mathsf{T}$ is positive definite, the first term in (56) is non-negative. This implies that, for any $\boldsymbol{k}$, the minimum is attained for $\boldsymbol{\theta} = \boldsymbol{\theta}^{\star|\boldsymbol{k}}$ (which annihilates the first summand in the objective function). Moreover, the second summand in (56) does not depend on $\boldsymbol{\theta}$. ∎

Summarizing the chain of implications presented so far, we conclude that for any solution $\boldsymbol{k}^\star$ of Problem 6 we can obtain a solution $(\boldsymbol{\theta}^\star, \boldsymbol{k}^\star)$ of Problem 4. Moreover, from $\boldsymbol{\theta}^\star$ we can easily obtain the solution of our original problem (Problem 3) by applying the modulus operation to $\boldsymbol{\theta}^\star$.

*2) Symmetries of Problem 6:* Because (49) and (56) are completely equivalent, they have the same symmetries. However, it is interesting to find the symmetries of Problem 6 directly. Notice that the reorganization of the terms made the term $C\boldsymbol{k}$ explicit. Because the $CP_\delta C^\mathsf{T}$ is positive definite, the only symmetries are described by the kernel of $C$. For any integer vector $\boldsymbol{q} \in \ker C$, this transformation does not change the value of the objective function:

$$\boldsymbol{k} \mapsto \boldsymbol{k} + \boldsymbol{q}, \quad \text{for } \boldsymbol{q} \in \ker C \cap \mathbb{Z}^m. \quad (70)$$

Recall that $C$ is a full-row-rank $\ell \times m$ matrix, where $\ell$ is the dimension of the cycle space. Its kernel $\ker C$ has thus dimension $m - \ell$, which is equal to $n$. As it happens, $A^\mathsf{T}$ is an orthogonal complement of $C$, so that it provides a base for its kernel. Therefore, any $\boldsymbol{q} \in \ker C \cap \mathbb{Z}^m$ can be written

as $q = A^\mathsf{T} p$, for some $p \in \mathbb{Z}^n$. Therefore, the symmetry can be written as

$$k \mapsto k + A^\mathsf{T} p, \qquad \text{for } p \in \mathbb{Z}^n, \tag{71}$$

which confirms the symmetry in (53).

### E. From $k$ towards a minimal parameterization $\gamma$

The cycle basis matrix $C$ is a "fat" $\ell \times m$ matrix, because the number of cycles $\ell$ is much less than the number of edges $m$. Therefore, there will be an infinite number of $k^\star$ such that the product $Ck^\star$ attains the minimum of (67). This infinite number is precisely described by the symmetry (71). Consequently, we will have an infinite number of optimal orientation estimates $\theta^{\star|k^\star}$. Fortunately, the next proposition assures that the infinite cardinality of solutions is an artifact created when passing from SO(2) to the reals. In, particular, we show that all vectors $k$ having the same product $Ck$ lead to orientation estimates $\theta^{\star|k}$ that differ by integer multiples of $2\pi$, hence corresponding to the same estimate, after an exponential map is applied.

**Proposition 12** (Equivalence of $k$ satisfying $Ck = \gamma$). *Consider a $k$ and the corresponding orientation estimate $\theta^{\star|k}$; then any $\bar{k}$ such that $C\bar{k} = Ck$ satisfies*

$$\theta^{\star|\bar{k}} = \theta^{\star|k} + 2\pi D(k - \bar{k}), \tag{72}$$

*where $D$ is a suitable integer matrix.*

*Proof:* Consider two regularization vectors $k_1$ and $k_2$ such that $Ck_1 = Ck_2 = \gamma$. Recall that $\theta^{\star|k_1} = (AP_\delta^{-1} A^\mathsf{T})^{-1} AP_\delta^{-1}(\check{\delta} - 2\pi k_1)$ and similarly $\theta^{\star|k_2} = (AP_\delta^{-1} A^\mathsf{T})^{-1} AP_\delta^{-1}(\check{\delta} - 2\pi k_2)$. Define $\delta^{\star|k_1} = A^\mathsf{T}\theta^{\star|k_1}$ and $\delta^{\star|k_2} = A^\mathsf{T}\theta^{\star|k_2}$. If the orientation of the first node is set to zero, $\delta^{\star|k_1}$ and $\delta^{\star|k_2}$ uniquely identify $\theta^{\star|k_1}$ and $\theta^{\star|k_2}$, since $\theta^{\star|k_i}$ can be rewritten as an integer-valued linear combination of $\delta^{\star|k_i}$, $i = 1, 2$; for instance, the orientation of node $j$ can be rewritten as $\theta_j^{\star|k_i} = \sum_{e \in p(0,j)} \lambda_e \delta_e^{\star|k_i}$, where $p(0, j)$ is the set of edges along a path connecting node 0 and node $j$ and $\lambda_e$ is $+1$ if the edges $e$ is traversed in forward direction along the path, $-1$ otherwise. In general, we have $\theta^{\star|k_i} = D\delta^{\star|k_i}, i = 1, 2$, where $D$ is an integer-valued matrix ($D$ is a left integer pseudoinverse of $A^\mathsf{T}$). Therefore, determining $\delta^{\star|k_i}$ is the same as determining $\theta^{\star|k_i}, i = 1, 2$.

The difference $\delta^{\star|k_2} - \delta^{\star|k_1}$ can be written as

$$\delta^{\star|k_2} - \delta^{\star|k_1} = 2\pi A^\mathsf{T}(AP_\delta^{-1} A^\mathsf{T})^{-1} AP_\delta^{-1}(k_1 - k_2) \tag{73}$$
(By Lemma 23)
$$= 2\pi(k_1 - k_2) + \tag{74}$$
$$-2\pi P_\delta C^\mathsf{T}(CP_\delta C^\mathsf{T})^{-1}(Ck_1 - Ck_2).$$

Since by assumption $Ck_1 = Ck_2$, the second term in (75) disappears, and one obtains

$$\delta^{\star|k_2} - \delta^{\star|k_1} = 2\pi(k_1 - k_2). \tag{75}$$

Therefore, elements of $\delta^{\star|k_1}$ and $\delta^{\star|k_2}$ only differ by multiples of $2\pi$; then, $\theta^{\star|k_2} - \theta^{\star|k_2} = D(\delta^{\star|k_2} - \delta^{\star|k_1}) = 2\pi D(k_1 - k_2)$, and since $D$ is an integer-valued matrix, also the elements of $\theta^{\star|k_1}$ and $\theta^{\star|k_2}$ only differ by multiples of $2\pi$, which concludes the proof. ∎

The previous result enables to solve Problem 6 directly on the slack variable

$$\gamma = Ck \in \mathbb{Z}^m. \tag{76}$$

In fact, all $k$ producing the same $\gamma = Ck$ are equivalent, in the sense specified in Proposition 12. Because $C$ is an integer matrix and $k$ is an integer vector, also $\gamma$ is an integer vector. The vector $\gamma$ clearly depends on the particular choice of the cycle basis matrix $C$.

The final formulation of the problem uses only $\gamma$.

**Problem 7** (Integer formulation of maximum-likelihood orientation estimation in cycle space). *Given the vector $\check{\delta} \in (-\pi, +\pi]^m$ and the diagonal positive definite matrix $P_\delta \in \mathbb{R}^{m \times m}$, and a cycle basis matrix $C \in \mathbb{Z}^{\ell \times m}$, find the set of minimizers $S^7 \subset \mathbb{Z}^\ell$ that satisfy*

$$S^7 = \arg\min_{\gamma \in \mathbb{Z}^\ell} \|\gamma - \hat{\gamma}\|^2_{P_\gamma^{-1}} \tag{77}$$

*with $\hat{\gamma} = \frac{1}{2\pi} C\check{\delta}$, and $P_\gamma = \frac{1}{4\pi^2} CP_\delta C^\mathsf{T}$.*

*1) From Problem 7 to Problem 6:* Given a $\gamma$, there is a simple way to compute a $k$ satisfying $Ck = \gamma$, assuming that the rows of $C$ are ordered appropriately as in (5).

**Lemma 13.** *Given a vector $\gamma \in \mathbb{Z}^\ell$ and a cycle basis matrix written as $C = (C_T \; C_L)$, an integer solution to $Ck = \gamma$ can be computed as*

$$k = \begin{pmatrix} 0_n \\ C_L^{-1} \gamma \end{pmatrix}. \tag{78}$$

*Proof:* From Liebchen [47, Lemma 3 and Theorem 7] it follows that $C_L$ is invertible and $\det(C_L) = \pm 1$. Moreover, because $C_L$ is an integer matrix with unitary determinant, necessarily $C_L^{-1}$ is itself integer (see Schrijver [48], or just think that the inverse is the adjoint matrix over the determinant). Therefore, $C_L^{-1}\gamma$ is an integer vector. Finally, we can show that $Ck = \gamma$ by inspection:

$$Ck = (C_T \; C_L)(0_n^\mathsf{T} \; (C_L^{-1}\gamma)^\mathsf{T})^\mathsf{T} = C_L(C_L^{-1}\gamma) = \gamma. \tag{79}$$
∎

We use the notation

$$C^\dagger = \begin{pmatrix} 0_n \\ C_L^{-1} \end{pmatrix} \tag{80}$$

to remark that $C^\dagger$ is a right (integer) pseudoinverse of $C$.

*2) Symmetries of Problem 7:* The objective function (77) is convex so it would be tempting to just say that there is only one minimum. However, we should be careful because the intuitions of convex optimization often fail in integer programming. For example, Figure 3 shows a case in which a convex objective function has two integer solutions.

What we can say is that this cannot happen for general data.

**Proposition 14.** $|S^7| = 1$ *with probability 1.*

*Proof:* The set $S^7$ is the set of minimizers of (77), which is an objective function of the form

$$\|\gamma - \mu\|^2_P, \tag{81}$$

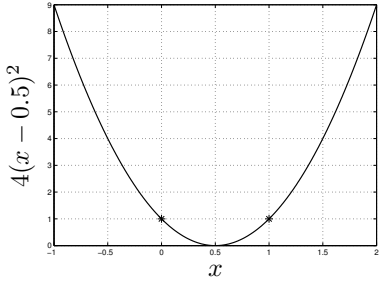

Fig. 3. Unconstrained quadratic integer program $\min_{x \in \mathbb{Z}} 4(x-0.5)^2$ in the scalar variable $x$. The optimal objective is $+1$ and is attained at the integer solutions $x_1^\star = 0$ and $x_2^\star = 1$.

with $\boldsymbol{P} \in \mathbb{R}^{\ell \times \ell}$ a positive definite matrix, and $\boldsymbol{\mu} \in \mathbb{R}^\ell$ a random variable which depends on the measurements, and can be seen to be a Gaussian vector. Consider the set of values $M \subset \mathbb{R}^\ell$ such that if $\boldsymbol{\mu} \in M$, then (81) has multiple solutions for $\boldsymbol{\gamma}$. Then it is easy to show that the set $M$ has measure 0 in $\mathbb{R}^\ell$. (For example, we can see how for an infinitesimal perturbation of the mean of the parabola in Figure 3, the solution set goes from $\{0,1\}$ to either $\{0\}$ or $\{1\}$.) For demonstrating this claim, consider a generic $\boldsymbol{\mu} \in M$. Since $\boldsymbol{\mu} \in M$ there exist $v \geq 2$ discrete values $\{\boldsymbol{\gamma}^1, \ldots, \boldsymbol{\gamma}^v\} \in \mathbb{Z}^\ell$ such that

$$\|\boldsymbol{\gamma}^1 - \boldsymbol{\mu}\|_{\boldsymbol{P}}^2 = \cdots = \|\boldsymbol{\gamma}^v - \boldsymbol{\mu}\|_{\boldsymbol{P}}^2. \tag{82}$$

If we fix $\{\boldsymbol{\gamma}^1, \ldots, \boldsymbol{\gamma}^v\}$, and take $\boldsymbol{\mu}$ as the independent variable, the constraints in (82) define an algebraic variety of dimension at most $\ell - 1$, which has measure 0 in $\mathbb{R}^\ell$. ∎

*F. Inception*

Let us link Problem 7 with the other problems presented so far. From a solution $\boldsymbol{\gamma}$ of Problem 7 we can find a solution $\boldsymbol{k}$ of Problem 6 using the formula (78). We call this map $\varphi_7^6$:

$$\varphi_7^6 : \mathbb{Z}^\ell \rightarrow \mathbb{Z}^m \tag{83}$$
$$\boldsymbol{\gamma} \mapsto \boldsymbol{C}^\dagger \boldsymbol{\gamma}. \tag{84}$$

However, there are much fewer $\boldsymbol{\gamma} \in \mathbb{Z}^\ell$ than $\boldsymbol{k} \in \mathbb{Z}^m$, therefore, using this map on $S^7$ we will not be able to cover all of $S^6$:

$$\varphi_7^6(S^7) \subsetneq S^6. \tag{85}$$

Likewise, further projecting down to using $\varphi_6^5$ we will still lose solutions:

$$\varphi_6^5 \circ \varphi_7^6(S^7) \subsetneq S^5. \tag{86}$$

Another step is not enough, as we still have

$$\varphi_5^4 \circ \varphi_6^5 \circ \varphi_7^6(S^7) \subsetneq S^4. \tag{87}$$

We have to go four levels deep. When we finally arrive back to Problem 3, we do obtain all solutions.

**Proposition 15** (Inception). $\varphi_4^3 \circ \varphi_5^4 \circ \varphi_6^5 \circ \varphi_7^6(S^7) = S^3$.

*Proof:* We first prove the relation $\varphi_4^3 \circ \varphi_5^4 \circ \varphi_6^5 \circ \varphi_7^6(S^7) \subseteq S^3$. Consider a $\boldsymbol{\gamma}^\star \in S^7$. By construction, $\boldsymbol{k}^\star = \varphi_7^6(\boldsymbol{\gamma}^\star)$ attains a minimum of Problem 6, since $\boldsymbol{\gamma}$ is only a slack variable that substitutes $\boldsymbol{Ck}$. Therefore, $\boldsymbol{k}^\star \in S^6$. Then, according to Proposition 11, $\varphi_6^5(\boldsymbol{k}^\star) = \varphi_6^5 \circ \varphi_7^6(\boldsymbol{\gamma}^\star)$ belongs to $S^5$, since $\varphi_6^5$ maps an optimal solution of Problem 6 to an optimal solution of Problem 5. Similarly, Proposition 8 and Proposition 9 imply that $\varphi_4^3 \circ \varphi_5^4 \circ \varphi_6^5 \circ \varphi_7^6(\boldsymbol{\gamma}^\star) \in S^3$.

Now, we want to prove $\varphi_4^3 \circ \varphi_5^4 \circ \varphi_6^5 \circ \varphi_7^6(S^7) \supseteq S^3$, that, together with the relation proved in the previous paragraph, demonstrates Proposition 15. Pick a $\check{\boldsymbol{\theta}}^\star \in S^3$. As shown in the proof of Proposition 8, $\boldsymbol{\theta}^\star = \check{\boldsymbol{\theta}}^\star$ also belongs to $S^4$, and it is such that (a) $\varphi_4^3(\boldsymbol{\theta}^\star) = \check{\boldsymbol{\theta}}^\star$. Now, we have already seen that $\boldsymbol{k}$ is only another way of writing the exponential map (see Section IV-B), therefore, if we choose $\boldsymbol{k}^\star = \lfloor(\pi \boldsymbol{1}_m - (\boldsymbol{A}^\mathsf{T} \boldsymbol{\theta}^\star - \check{\boldsymbol{\delta}}))/2\pi\rfloor$ (compare with equations (7) and (49)), then the pair $(\boldsymbol{\theta}^\star, \boldsymbol{k}^\star)$ solves Problem 5; moreover, $(\boldsymbol{\theta}^\star, \boldsymbol{k}^\star)$ is such that $\varphi_5^4(\boldsymbol{\theta}^\star, \boldsymbol{k}^\star) = \boldsymbol{\theta}^\star$, that, using (a), implies (b) $\varphi_4^3 \circ \varphi_5^4(\boldsymbol{\theta}^\star, \boldsymbol{k}^\star) = \check{\boldsymbol{\theta}}^\star$. According to Proposition 11, if $(\boldsymbol{\theta}^\star, \boldsymbol{k}^\star) \in S^5$, then $\boldsymbol{k}^\star \in S^6$, and then, from (b), we get (c) $\varphi_4^3 \circ \varphi_5^4 \circ \varphi_6^5(\boldsymbol{k}^\star) = \check{\boldsymbol{\theta}}^\star$. Finally, if $\boldsymbol{k}^\star$ is an optimal solution for Problem 6, we can pick $\boldsymbol{\gamma}^\star = \boldsymbol{Ck}^\star$ and guarantee that $\boldsymbol{\gamma}^\star$ is an optimal solution for Problem 7 (there is only a change of variables between the two problems), i.e., $\boldsymbol{\gamma}^\star \in S^7$. Concluding, for a given $\check{\boldsymbol{\theta}}^\star \in S^3$, we found a $\boldsymbol{\gamma}^\star \in S^7$, that is such that $\check{\boldsymbol{\theta}}^\star = \varphi_4^3 \circ \varphi_5^4 \circ \varphi_6^5 \circ \varphi_7^6(\boldsymbol{\gamma}^\star)$. ∎

*G. Conclusions*

The following theorem, whose proof is a direct consequence of Proposition 14 and 15, summarizes what we have learned so far.

**Theorem 16** (Mixed-integer maximum likelihood estimation on $SO(2)$). *The solution of Problem 3 is almost surely unique, and, given the solution $\boldsymbol{\gamma}^\star$ of*

$$\boldsymbol{\gamma}^\star = \arg\min_{\boldsymbol{\gamma} \in \mathbb{Z}^\ell} \left\|\boldsymbol{\gamma} - \tfrac{1}{2\pi}\boldsymbol{C}\check{\boldsymbol{\delta}}\right\|_{\boldsymbol{P}_{\boldsymbol{\gamma}}^{-1}}^2, \tag{88}$$

*can be computed as*

$$\check{\boldsymbol{\theta}}^\star = \left\langle (\boldsymbol{A}\boldsymbol{P}_\delta^{-1}\boldsymbol{A}^\mathsf{T})^{-1}\boldsymbol{A}\boldsymbol{P}_\delta^{-1}\left(\check{\boldsymbol{\delta}} - 2\pi\boldsymbol{C}^\dagger\boldsymbol{\gamma}^\star\right)\right\rangle_{2\pi}, \tag{89}$$

*with* $\boldsymbol{C}^\dagger \doteq \begin{pmatrix} \boldsymbol{0}_n \\ \boldsymbol{C}_L^{-1} \end{pmatrix}$. □

It is easy to see that (89) is a closed-form expression for the mapping $\varphi_4^3 \circ \varphi_5^4 \circ \varphi_6^5 \circ \varphi_7^6(\boldsymbol{\gamma}^\star)$, hence the result is a more explicit version of Proposition 15.

In contrast with iterative optimization techniques, the proposed computation of $\check{\boldsymbol{\theta}}^{\star|\boldsymbol{\gamma}^\star}$ does not suffer from local minima, assuming that we are able to compute $\boldsymbol{\gamma}^\star$.

Although this seems a good conclusion, we are not yet done with our estimation problem for two main reasons.

First, Problem 7 is NP-hard [36]. Several algorithms have been proposed in literature to solve integer quadratic programming (see, e.g., [39], [49], [50] and references therein); however, for the hardness of the problem, one cannot expect to solve exactly and *quickly* large-scale problems; for example, Chang and Golub [51] report computational times above 10 seconds in problem instances with less than 50 variables. Similar numerical results are reported by Jazaeri *et al.* [50].

Second, as explained in the next section, computing only the maximum likelihood estimate does not guarantee to have an accurate orientation estimate.

## V. LIMITATIONS OF THE MAXIMUM LIKELIHOOD ESTIMATE

Section IV has shown that the maximum likelihood optimization problem on the orientations, which lie on the product manifold $\mathrm{SO}(2)^n$, is equivalent to an optimization problem in the variable $\boldsymbol{\gamma}$, which lies on the integers lattice $\mathbb{Z}^\ell$. The variable $\boldsymbol{\gamma}$ can be thought of an alternative reparametrization of the problem.

This section shows that, if the maximum likelihood solution $\boldsymbol{\gamma}^\star$ is different than the "true" value $\boldsymbol{\gamma}^\circ$, then there is a bias in the corresponding estimate of $\boldsymbol{\theta}$ (Lemma 17). This bias raises the mean square estimation error of the maximum likelihood estimate. Section V-B discusses a simple practical example for a circle graph. In the next section we will describe an algorithm that, instead of estimating only one value for $\boldsymbol{\gamma}$, returns multiple hypotheses for $\boldsymbol{\gamma}$ which contain $\boldsymbol{\gamma}^\circ$ with a given probability, so that we will be able to prove that at least one of the hypotheses gives a small estimation error.

### A. Distribution of maximum-likelihood estimate

To define the "true" value of $\boldsymbol{\gamma}$ we rewrite the measurement model as
$$\check{\boldsymbol{\delta}} = \langle \boldsymbol{A}^\mathsf{T}\check{\boldsymbol{\theta}}^\circ + \boldsymbol{\epsilon}\rangle_{2\pi}, \quad (90)$$
where $\check{\boldsymbol{\theta}}^\circ$ is the "true" value of the nodes orientations. From (7) the measurement model can be written as
$$\check{\boldsymbol{\delta}} = \boldsymbol{A}^\mathsf{T}\check{\boldsymbol{\theta}}^\circ + 2\boldsymbol{k}^\circ \pi + \boldsymbol{\epsilon}, \quad (91)$$
where $\boldsymbol{k}^\circ$ is the "true" regularization vector, and has the expression $\boldsymbol{k}^\circ \doteq \lfloor (\pi \boldsymbol{1}_m - \boldsymbol{A}^\mathsf{T}\check{\boldsymbol{\theta}}^\circ - \boldsymbol{\epsilon})/2\pi \rfloor$. Define the "true" value of $\boldsymbol{\gamma}$ as
$$\boldsymbol{\gamma}^\circ \doteq \boldsymbol{C}\boldsymbol{k}^\circ. \quad (92)$$
For a given $\boldsymbol{\gamma}$, define the *real-valued estimate* $\boldsymbol{\theta}^{\star|\boldsymbol{\gamma}}$ as
$$\boldsymbol{\theta}^{\star|\boldsymbol{\gamma}} \doteq (\boldsymbol{A}\boldsymbol{P}_\delta^{-1}\boldsymbol{A}^\mathsf{T})^{-1}\boldsymbol{A}\boldsymbol{P}_\delta^{-1}(\check{\boldsymbol{\delta}} - 2\pi\boldsymbol{C}^\dagger\boldsymbol{\gamma}) \in \mathbb{R}^n. \quad (93)$$
and the corresponding (wrapped) estimate $\check{\boldsymbol{\theta}}^{\star|\boldsymbol{\gamma}}$ as:
$$\check{\boldsymbol{\theta}}^{\star|\boldsymbol{\gamma}} \doteq \langle \boldsymbol{\theta}^{\star|\boldsymbol{\gamma}}\rangle_{2\pi} \in (-\pi, +\pi]^n. \quad (94)$$
We call $\boldsymbol{\theta}^{\star|\boldsymbol{\gamma}^\star}$ the *real-valued maximum likelihood estimate* (it is $\boldsymbol{\theta}^{\star|\boldsymbol{\gamma}}$ computed for $\boldsymbol{\gamma}^\star$). Comparing with (89), the reader can notice that *real-valued maximum likelihood estimate* is simply the maximum likelihood orientation estimate before applying the exponential map. We can give a full characterization of the distribution of the real-valued maximum likelihood estimator.

**Lemma 17.** *The real-valued maximum likelihood estimator $\boldsymbol{\theta}^{\star|\boldsymbol{\gamma}^\star}$ can be written as*
$$\begin{aligned}
\boldsymbol{\theta}^{\star|\boldsymbol{\gamma}^\star} &= \check{\boldsymbol{\theta}}^\circ + & (95)\\
&\quad 2\pi\boldsymbol{p} + & (96)\\
&\quad (\boldsymbol{A}\boldsymbol{P}_\delta^{-1}\boldsymbol{A}^\mathsf{T})^{-1}\boldsymbol{A}\boldsymbol{P}_\delta^{-1}\boldsymbol{\epsilon} + & (97)\\
&\quad 2\pi(\boldsymbol{A}\boldsymbol{P}_\delta^{-1}\boldsymbol{A}^\mathsf{T})^{-1}\boldsymbol{A}\boldsymbol{P}_\delta^{-1}\boldsymbol{C}^\dagger(\boldsymbol{\gamma}^\circ - \boldsymbol{\gamma}^\star), & (98)
\end{aligned}$$
*where the term* (95) *is the true value of the orientations; the term* (96) *contains some integer vector $\boldsymbol{p}$ and has no effect once the exponential map is applied to $\boldsymbol{\theta}^{\star|\boldsymbol{\gamma}^\star}$; the term* (97) *contains the noise which would appear even if the problem were linear; and, finally, the term* (98) *contains an additional bias, which is proportional to the mismatch between $\boldsymbol{\gamma}^\star$ and $\boldsymbol{\gamma}^\circ$. In particular, if $\boldsymbol{\gamma}^\star = \boldsymbol{\gamma}^\circ$, then the vector $\boldsymbol{\theta}^{\star|\boldsymbol{\gamma}}$ is Normally distributed with mean $\check{\boldsymbol{\theta}}^\circ + 2\pi\boldsymbol{p}$ and covariance matrix $(\boldsymbol{A}\boldsymbol{P}_\delta^{-1}\boldsymbol{A}^\mathsf{T})^{-1}$.*

*Proof:* Substitute $\check{\boldsymbol{\delta}}$ from (91) in the expression of the real-valued estimator to obtain
$$\boldsymbol{\theta}^{\star|\boldsymbol{\gamma}} = (\boldsymbol{A}\boldsymbol{P}_\delta^{-1}\boldsymbol{A}^\mathsf{T})^{-1}\boldsymbol{A}\boldsymbol{P}_\delta^{-1}\left(\boldsymbol{A}^\mathsf{T}\check{\boldsymbol{\theta}}^\circ + \boldsymbol{\epsilon} + 2\pi(\boldsymbol{k}^\circ - \boldsymbol{C}^\dagger\boldsymbol{\gamma})\right). \quad (99)$$
Rewrite $\boldsymbol{k}^\circ = \boldsymbol{C}^\dagger\boldsymbol{\gamma}^\circ + \boldsymbol{k}^\perp$, as to separate $\boldsymbol{k}^\circ$ into two vectors: the first is $\boldsymbol{C}^\dagger\boldsymbol{\gamma}^\circ$, which, by construction, satisfies $\boldsymbol{C}(\boldsymbol{C}^\dagger\boldsymbol{\gamma}^\circ) = \boldsymbol{\gamma}^\circ$. The second belongs to the kernel of $\boldsymbol{C}$, and then satisfies $\boldsymbol{C}\boldsymbol{k}^\perp = \boldsymbol{0}_\ell$. Note that $\boldsymbol{k}^\perp$ is an integer vector, since both $\boldsymbol{k}^\circ$ and $\boldsymbol{C}^\dagger\boldsymbol{\gamma}^\circ$ are integer vectors by construction. According to the result in Proposition 12, the term $\boldsymbol{k}^\perp$ only adds multiples of $2\pi$ to the estimate
$$\begin{aligned}
\boldsymbol{\theta}^{\star|\boldsymbol{\gamma}} &= (\boldsymbol{A}\boldsymbol{P}_\delta^{-1}\boldsymbol{A}^\mathsf{T})^{-1}\boldsymbol{A}\boldsymbol{P}_\delta^{-1} \times \\
&\quad \left(\boldsymbol{A}^\mathsf{T}\check{\boldsymbol{\theta}}^\circ + \boldsymbol{\epsilon} + 2\pi\boldsymbol{C}^\dagger(\boldsymbol{\gamma}^\circ - \boldsymbol{\gamma})\right) + 2\pi\boldsymbol{p}, \quad (100)
\end{aligned}$$
with $\boldsymbol{p} = \boldsymbol{D}\boldsymbol{k}^\perp$, for some integer matrix $\boldsymbol{D}$. Multiplying the parentheses gives the desired result. ∎

One consequence of this result is that the maximum-likelihood estimate of $\boldsymbol{\theta}$ computed from $\boldsymbol{\gamma}^\star$ is not necessarily the one with the minimum estimation error, because one cannot guarantee in general that the optimal solution $\boldsymbol{\gamma}^\star$ of Problem 7 is such that $\boldsymbol{\gamma}^\star = \boldsymbol{\gamma}^\circ$. Consequently, in order to attain a small estimation error, one should look for $\boldsymbol{\gamma}^\circ$, instead of simply computing $\boldsymbol{\gamma}^\star$.

### B. A simple example

Consider the pose graph of Figure 4. The robot traveled along a circle and came back to the starting position $(0,0)$ after 18 time steps. Robot poses at each time step are plotted as black triangles (rotated according to the true robot orientations $\check{\boldsymbol{\theta}}^\circ$). The positions are shown only for representation purposes, while we are only interested in estimating the orientations from the relative orientation measurements. We build the relative orientation measurements as $\check{\boldsymbol{\delta}} = \langle \boldsymbol{A}^\mathsf{T}\check{\boldsymbol{\theta}}^\circ + \boldsymbol{\epsilon}\rangle_{2\pi}$, where $\boldsymbol{A}^\mathsf{T}$ is the incidence matrix of the graph, and $\boldsymbol{\epsilon}$ is the noise. For sake of repeatability we fixed the noise on each edge to $0.2$ rad. Now, we want to compare the real-valued maximum likelihood orientation estimate $\boldsymbol{\theta}^{\star|\boldsymbol{\gamma}^\star}$ and the real-valued orientation estimate $\boldsymbol{\theta}^{\star|\boldsymbol{\gamma}^\circ}$, obtained using $\boldsymbol{\gamma}^\circ$. In this toy example, $\boldsymbol{\gamma}^\star$ and $\boldsymbol{\gamma}^\circ$ are scalar quantities, since the cycle basis matrix is a vector, i.e., $\boldsymbol{C} = \boldsymbol{1}_m^\mathsf{T}$. In a real problem, one cannot compute $\boldsymbol{\gamma}^\circ$ since it is a function of the noise; however, in simulation, we can simply apply the definition $\boldsymbol{\gamma}^\circ \doteq \boldsymbol{C} \lfloor (\pi\boldsymbol{1}_m - \boldsymbol{A}^\mathsf{T}\check{\boldsymbol{\theta}}^\circ - \boldsymbol{\epsilon})/2\pi \rfloor$ (we obtain $\boldsymbol{\gamma}^\circ = 1$ in our example). The vector $\boldsymbol{\gamma}^\star$ can be instead computed as solution of Problem 7. In this case the problem is solved over a scalar unknown, and its solution corresponds to rounding $\hat{\gamma}$ to the closest integer; in the example we obtain $\boldsymbol{\gamma}^\star = 2$. Then, we use the expression (89) to obtain $\boldsymbol{\theta}^{\star|\boldsymbol{\gamma}^\star}$ and $\boldsymbol{\theta}^{\star|\boldsymbol{\gamma}^\circ}$. Let us now compute the likelihood of these vectors using the cost (35)

(lower cost corresponds to higher likelihood). According to the property (11) the cost is the same for the real-valued estimate and for the corresponding wrapped estimate. As expected, $\boldsymbol{\theta}^{\star|\gamma^\star}$, which is the real-valued maximum likelihood estimate, gives the smallest value of the cost function (35) (the value is $0.4$ in our example). Computing the cost function (35) in $\boldsymbol{\theta}^{\star|\gamma^\circ}$ we obtain a higher cost (i.e., a lower likelihood): $0.72$ in our example. Everything suggests that $\boldsymbol{\theta}^{\star|\gamma^\star}$ is a better estimate of $\check{\boldsymbol{\theta}}^\circ$ is compared with $\boldsymbol{\theta}^{\star|\gamma^\circ}$. Figure 4, however, tells another story: the estimate $\boldsymbol{\theta}^{\star|\gamma^\circ}$ (represented as blue arrows) is a very good estimate of $\check{\boldsymbol{\theta}}^\circ$, while the maximum likelihood estimate $\boldsymbol{\theta}^{\star|\gamma^\star}$ (represented as red arrows) is far from the actual orientations.

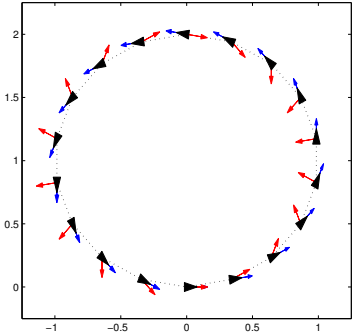

Fig. 4. A counterexample to the use of the maximum likelihood orientation estimate. The figure shows a simple pose graph in which the robot is travelling along a circle and is coming back to the starting position $(0,0)$ after 18 time steps. Robot poses are shows as black triangles. The real-valued maximum likelihood orientation estimate $\boldsymbol{\theta}^{\star|\gamma^\star}$ is represented as red arrows. The blue arrows are the orientation estimate $\boldsymbol{\theta}^{\star|\gamma^\circ}$ obtained from $\boldsymbol{\gamma}^\circ$.

One may argue that this problem arises only for the particular realization of the noise we considered. However, this is not the case: the same issue appears whenever the sum of the orientation noises along the cycle is larger (in absolute value) than $\pi$, and this is going to happen also in presence of Gaussian noise (i.e., it cannot be disregarded as an improbable event). In our toy example we can explicitly derive the probability of this event. The probability that the sum of the noises along the cycle $\boldsymbol{c}$ is bigger than $\pi$ is

$$\mathbb{P}\left(\left|\sum_{(i,j)\in\boldsymbol{c}}\epsilon_{ij}\right| > \pi\right) = \mathbb{P}(|\epsilon_{\text{cycle}}| > \pi), \quad (101)$$

where $\epsilon_{\text{cycle}}$ is a Gaussian random variable with variance $\sigma^2_{\text{cycle}} = \sum_{(i,j)\in\boldsymbol{c}} \sigma^2_{ij}$. Using standard results on Gaussian distributions we conclude

$$\mathbb{P}(|\epsilon_{\text{cycle}}| > \pi) = 1 - \mathbf{erf}\left(\frac{\pi}{\sigma_{\text{cycle}}\sqrt{2}}\right), \quad (102)$$

where $\mathbf{erf}$ denotes the *error function*. The relation (102) underlines how the probability of "wraparound" (i.e., mismatch between $\boldsymbol{\gamma}^\star$ and $\boldsymbol{\gamma}^\circ$) increases with $\sigma_{\text{cycle}}$, which, in turns, depends on the *length* of the cycle (number of edges belonging to the cycle) and on the uncertainty of the corresponding measurements.

*C. Interpretation*

These results are not surprising, if we realize that the problem is nonlinear. In fact, only in linear Gaussian models one can guarantee that the maximum likelihood estimator is unbiased and achieves the minimum estimation error, while it is not necessary true in the general nonlinear case.

Ultimately, we are led to conclude that in the case of high sensor noise, orientation estimation is a problem whose structure is not particularly suited for a maximum-likelihood formalization. More abstractly, for a suitable distance function "**dist**" on the manifold, minimizing the *data error* [52], i.e., $\mathbf{dist}(\boldsymbol{A}^\top\boldsymbol{\theta}, \check{\boldsymbol{\delta}})$, does not ensure to have a low value for the *estimation error*, which is $\mathbf{dist}(\boldsymbol{\theta}, \check{\boldsymbol{\theta}}^\circ)$.

Consequently, rather than finding the maximum likelihood estimate for $\boldsymbol{\gamma}$, we devised an algorithm, presented in the next section, that tries to find the value $\boldsymbol{\gamma}^\circ$, or, more precisely, it finds a set of integer vectors that is guaranteed to contain $\boldsymbol{\gamma}^\circ$ with a given probability.

## VI. A Multi-hypothesis Estimator for $\boldsymbol{\gamma}^\circ$

This section describes an algorithm that we call INTEGER-SCREENING which is able to find a set $\Gamma$ of integer vectors that contains $\boldsymbol{\gamma}^\circ$ with a desired probability.

Section VI-A describes how to derive an estimator for $\boldsymbol{\gamma}^\circ$ which allows building a confidence set.

Section VI-B describes the INTEGER-SCREENING algorithm, which builds a set of integer vectors $\Gamma$ that contains $\boldsymbol{\gamma}^\circ$ with desired probability.

Section VI-C discusses the influence of the cycle basis matrix in the construction of $\Gamma$ and formally proves that the minimum cycle basis matrix MCB is the optimal choice that minimizes the expected size of the set $\Gamma$.

*A. An estimator of $\boldsymbol{\gamma}^\circ$*

From the knowledge of a cycle basis matrix $\boldsymbol{C}$, the measurements $\check{\boldsymbol{\delta}}$, and the covariance matrix $\boldsymbol{P}_\delta$ we can design an estimator $\hat{\boldsymbol{\gamma}}$ for the unknown integer vector $\boldsymbol{\gamma}^\circ$.

**Proposition 18.** *The real-valued vector*

$$\hat{\boldsymbol{\gamma}} \doteq \tfrac{1}{2\pi}\boldsymbol{C}\check{\boldsymbol{\delta}} \quad (103)$$

*is a Normally distributed estimator of $\boldsymbol{\gamma}^\circ$, with covariance matrix*

$$\boldsymbol{P}_\gamma \doteq \tfrac{1}{4\pi^2}\boldsymbol{C}\boldsymbol{P}_\delta\boldsymbol{C}^\top. \quad (104)$$

*Proof:* Multiply both members of (91) by $\boldsymbol{C}$ to obtain

$$\boldsymbol{C}\check{\boldsymbol{\delta}} = \boldsymbol{C}(\boldsymbol{A}^\top\check{\boldsymbol{\theta}}^\circ + 2\pi\boldsymbol{k}^\circ + \boldsymbol{\epsilon}). \quad (105)$$

By Lemma 1, the term $\boldsymbol{C}\boldsymbol{A}^\top$ is equal to zero. Reordering the remaining terms, we get $\boldsymbol{C}\boldsymbol{k}^\circ = \tfrac{1}{2\pi}\boldsymbol{C}\check{\boldsymbol{\delta}} - \tfrac{1}{2\pi}\boldsymbol{C}\boldsymbol{\epsilon}$. Using the definition $\boldsymbol{\gamma}^\circ = \boldsymbol{C}\boldsymbol{k}^\circ$ given in Section V, this implies $\boldsymbol{\gamma}^\circ = \tfrac{1}{2\pi}\boldsymbol{C}\check{\boldsymbol{\delta}} - \tfrac{1}{2\pi}\boldsymbol{C}\boldsymbol{\epsilon}$, from which the thesis follows. ∎

The availability of the estimator of Proposition 18 enables the computation of the set $\Gamma$, as described in the following section.

## B. The INTEGER-SCREENING algorithm

The INTEGER-SCREENING algorithm (Algorithm 1) computes a finite set of integer vectors that contains $\boldsymbol{\gamma}^\circ$ with a user-specified probability. The algorithm is based on two simple ideas: *marginalization* and *conditioning*. We use *marginalization* to exclude non-plausible values for the elements of $\boldsymbol{\gamma}^\circ$. Since $\hat{\boldsymbol{\gamma}}$ is Normally distributed, i.e., $\hat{\boldsymbol{\gamma}} \sim \mathcal{N}(\boldsymbol{\gamma}^\circ, \boldsymbol{P}_\gamma)$, also the marginal distribution of the $i$-th element $\hat{\gamma}_i$ is a Gaussian with mean $\gamma_i^\circ$. Therefore we easily derive a confidence interval for the single element, based on a given confidence level. If the interval contains only one integer, then we can assign that value to the element with the specified confidence. Once we are sure of the value of one element, say $\gamma_i^\circ = u_i$, we use *conditioning* to reduce the plausible values of the others, by conditioning on $\gamma_i^\circ = u_i$. These two ideas suggest an iterative algorithm that looks for elements of $\boldsymbol{\gamma}^\circ$ that can be determined unambiguously, and then uses those constraints to further shrink the uncertainty on the remaining elements.

The input to Algorithm 1 consists of the real vector $\hat{\boldsymbol{\gamma}}$, the positive semidefinite matrix $\boldsymbol{P}_\gamma$, and the confidence level $\alpha$. It is assumed that $\hat{\boldsymbol{\gamma}}$ is a Normally distributed estimator of $\boldsymbol{\gamma}^\circ \in \mathbb{Z}^\ell$ and that $\boldsymbol{P}_\gamma$ is its covariance. The output of the algorithm is a set $\Gamma$ of integer vectors that has probability no smaller than $\alpha$ of containing $\boldsymbol{\gamma}^\circ$; the probability bound is derived in Proposition 19 and Corollary 20.

Throughout the execution, the set $\mathcal{U}^{(k)}$ contains the indices of the elements of $\boldsymbol{\gamma}^\circ$ that are uniquely identified at iteration $k$, and, conversely, the set $\mathcal{R}^{(k)}$ that contains the indices that are still ambiguous. At the beginning (line 14), $\mathcal{R}^{(k)} = \{1, \ldots, \ell\}$ as no element has been identified.

We use the notation $\boldsymbol{\gamma}^\circ_{\mathcal{U}^{(k)}}$ to indicate the subvector of $\boldsymbol{\gamma}^\circ$ at the indices given by $\mathcal{U}^{(k)}$, and the notation $\boldsymbol{\gamma}^\circ_{\mathcal{R}^{(k)}}$ to denote the elements of $\boldsymbol{\gamma}^\circ$ at the indices given by $\mathcal{R}^{(k)}$. The algorithm updates two variables $\boldsymbol{\zeta}_{\mathcal{R}^{(k)}}$ and $\boldsymbol{P}_{\mathcal{R}^{(k)}}$. The invariant that holds is that

$$\boldsymbol{\zeta}_{\mathcal{R}^{(k)}} \sim \mathcal{N}(\boldsymbol{\gamma}^\circ_{\mathcal{R}^{(k)}}, \boldsymbol{P}_{\mathcal{R}^{(k)}}), \qquad (106)$$

i.e., they describe a Normally distributed estimator of the elements $\boldsymbol{\gamma}^\circ_{\mathcal{R}^{(k)}}$ that have not been identified yet. The invariant holds at the beginning as the variables are initialized to $\boldsymbol{\zeta}_{\mathcal{R}^{(k)}} = \hat{\boldsymbol{\gamma}}$ and $\boldsymbol{P}_{\mathcal{R}^{(k)}} = \boldsymbol{P}_\gamma$ (line 15).

At a generic iteration $k$, the algorithm computes the confidence set for each $\gamma_i^\circ$, $i \in \mathcal{R}^{(k)}$. Since $\boldsymbol{\zeta}_{\mathcal{R}^{(k)}}$ is Normally distributed, also each component $\zeta_i^{(k)}$ is Normally distributed with mean $\gamma_i^\circ$ and variance given by the $i$-th element of the covariance matrix $\boldsymbol{P}_{ii}^{(k)}$. Therefore, with probability $\eta$, it holds that

$$\gamma_i^\circ \in \left[\, \zeta_i^{(k)} - b, \zeta_i^{(k)} + b \,\right] \doteq \mathcal{I}_i^{(k)}, \qquad (107)$$

with $b = \sqrt{\boldsymbol{P}_{ii}^{(k)} \chi^2_{1,\eta}}$ (lines 21–22). Note that the confidence interval depends on $\eta = \alpha^{\frac{1}{\ell}}$. The relation between $\alpha$ and $\eta$ is justified by Proposition 19.

Then, the algorithm computes all the integers within the interval $\mathcal{I}_i^{(k)}$, obtaining $\Gamma_i^{(k)} = \mathcal{I}_i^{(k)} \cap \mathbb{Z}$ (line 23). If the set $\Gamma_i^{(k)}$ contains a single integer, say $u_i$, then with probability $\eta$ it holds that $\gamma_i^\circ = u_i$, and the algorithm adds the index $i$ to the set of uniquely determined elements $\mathcal{U}^{(k)}$ (line 26).

After checking all sets $\Gamma_i^{(k)}$, $i \in \mathcal{R}^{(k)}$ (line 27), we have the set $\mathcal{U}^{(k)}$, that contains all the elements in $\boldsymbol{\gamma}^\circ$ that we uniquely determined at the current iteration. Clearly, these indices can be removed from the ones that are still ambiguous (line 34). Moreover, we can exploit this information to infer the value of the remaining elements of $\boldsymbol{\gamma}^\circ$. In particular, we can compute the probability density $\mathbb{P}(\boldsymbol{\zeta}_{\mathcal{R}^{(k)}} | \boldsymbol{\zeta}_{\mathcal{U}^{(k)}} = \Gamma_{\mathcal{U}^{(k)}})$, i.e., the conditional probability density of the elements that are still ambiguous, given the values that we found for the elements in $\mathcal{U}^{(k)}$ (line 38). Since the original density is a Gaussian, also the conditional density is a Gaussian, with mean $\boldsymbol{\zeta}_{\mathcal{R}^{(k+1)}}$ and covariance $\boldsymbol{P}_{\mathcal{R}^{(k+1)}}$. Therefore, at the end of the iteration, we have a unique value for the elements in $\mathcal{U}^{(k)}$ and a probabilistic description (i.e., mean and covariance) of the elements in $\mathcal{R}^{(k+1)}$. Since the conditioning may have shrunk the uncertainty on some element, we proceed to the next iteration. If the set $\mathcal{U}^{(k)}$ is empty, it means that we are not able to make any progress and the loop exits (line 36). Otherwise, the algorithm shrinks the uncertainty by conditioning the current estimate on the information $\gamma_i^\circ = u_i$ for all $i \in \mathcal{U}^{(k)}$, and proceeds to the next iteration. Notice that when conditioning on some component of $\boldsymbol{\gamma}^\circ$ we reduce the size of the mean vector and the covariance matrix. At iteration $k$, it holds $\boldsymbol{\zeta}_{\mathcal{R}^{(k)}} \in \mathbb{R}^{|\mathcal{R}^{(k)}|}$ and $\boldsymbol{P}_{\mathcal{R}^{(k)}} \in \mathbb{R}^{|\mathcal{R}^{(k)}| \times |\mathcal{R}^{(k)}|}$. The algorithm performs at most $K \leq \ell$ iterations because at each iteration, at least one additional element of $\boldsymbol{\gamma}^\circ \in \mathbb{R}^\ell$ is determined.

After the algorithm stops we have a collection of confidence sets $\Gamma_i^{(K)} \subset \mathbb{Z}$, $i \in \{1, \ldots, \ell\}$. If the admissible elements $\gamma_i^\circ$ has to belong to $\Gamma_i^{(K)}$, then it must hold that

$$\boldsymbol{\gamma}^\circ \in \Gamma_1^{(K)} \times \Gamma_2^{(K)} \times \ldots \times \Gamma_\ell^{(K)} \doteq \Gamma, \qquad (108)$$

where $\times$ denotes the Cartesian product of sets (line 40). Corollary 20 bounds the probability of $\boldsymbol{\gamma}^\circ \in \Gamma$.

**Proposition 19** (Probability bounds for the confidence intervals in INTEGER-SCREENING). *At each iteration $k$ of Algorithm 1, the computed intervals $\mathcal{I}_i^{(k)}$, $i = \{1, \ldots, \ell\}$ are such that all elements of $\boldsymbol{\gamma}^\circ$ belong to the corresponding intervals with probability at least $\alpha$:*

$$\mathbb{P}\left(\gamma_1^\circ \in \mathcal{I}_1^{(k)} \wedge \ldots \wedge \gamma_\ell^\circ \in \mathcal{I}_\ell^{(k)}\right) \geq \alpha. \qquad (109)$$

*Proof:* The proof is based on the technical result of Lemma 24, reported in Appendix. Let us first manipulate the claim (109) to make it closer to the result (130). In Lemma 24 we discuss the probability of the random variable belonging to an interval centered on the mean, in (109) we have the probability of the mean belonging to the interval obtained from the random estimator. This is a minor problem, since we can "center" the confidence interval on the mean, i.e., we can write the event

$$\gamma_i^\circ \in \left[\zeta_i^{(k)} - \sqrt{\boldsymbol{P}_{ii}^{(k)} \chi^2_{1,\eta}},\ \zeta_i^{(k)} + \sqrt{\boldsymbol{P}_{ii}^{(k)} \chi^2_{1,\eta}}\right] \qquad (110)$$

in the equivalent form

$$\zeta_i^{(k)} \in \left[\gamma_i^\circ - \sqrt{\boldsymbol{P}_{ii}^{(k)} \chi^2_{1,\eta}},\ \gamma_i^\circ + \sqrt{\boldsymbol{P}_{ii}^{(k)} \chi^2_{1,\eta}}\right]. \qquad (111)$$

Call this centered confidence intervals $\bar{\mathcal{I}}_i$. Therefore, we rewrite the claim (109) as

$$\mathbb{P}\Big(\bigwedge_i \gamma_i^\circ \in \mathcal{I}_i^{(k)}\Big) = \mathbb{P}\Big(\bigwedge_i \zeta_i^{(k)} \in \bar{\mathcal{I}}_i^{(k)}\Big). \quad (112)$$

Before starting the analysis of the algorithm, note that, at the end of each iteration $k = \{1, \ldots, K\}$, the following relation holds between the sets of indices $\mathcal{U}^{(k)}$ and $\mathcal{R}^{(k)}$:

$$\mathcal{U}^{(1)} \cup \cdots \cup \mathcal{U}^{(k)} \cup \mathcal{R}^{(k+1)} = \{1, \ldots, \ell\}. \quad (113)$$

This relation is a consequence of the definition of $\mathcal{R}^{(k+1)}$, which contains the indices of the elements that have not been uniquely determined at iteration $k$. Moreover, by construction, a given index can only be contained in one of the sets $\mathcal{U}^{(1)}, \ldots, \mathcal{U}^{(k)}, \mathcal{R}^{(k+1)}$ which implies another useful relation (valid at any iteration $k$):

$$|\mathcal{U}^{(1)}| + \cdots + |\mathcal{U}^{(k)}| + |\mathcal{R}^{(k+1)}| = \ell. \quad (114)$$

Let now start our analysis from the first iteration of Algorithm 1. The quantity $\boldsymbol{\zeta}_{\mathcal{R}^{(1)}}$ has been initialized to $\hat{\boldsymbol{\gamma}}$, therefore it is a random vector with mean value $\boldsymbol{\gamma}^\circ$ and covariance matrix $\boldsymbol{P}_\gamma$. For this reason we can use Lemma 24 to conclude that

$$p^{(1)} \doteq \mathbb{P}\left(\zeta_1^{(1)} \in \bar{\mathcal{I}}_1^{(1)} \wedge \ldots \wedge \zeta_\ell^{(1)} \in \bar{\mathcal{I}}_\ell^{(1)}\right) \geq \eta^\ell = \alpha \quad (115)$$

which proves the claim of the proposition for the first iteration. Before proceeding to the next iteration, we notice that Lemma 24 can be also applied to a subset of variables $\boldsymbol{\zeta}_{\mathcal{R}^{(1)}}$ (since the overall vector is Normally distributed, also every subvector is Normally distributed). In particular, it is interesting to notice that

$$\mathbb{P}\Big(\bigwedge_{i \in \mathcal{U}^{(1)}} \zeta_i^{(1)} \in \bar{\mathcal{I}}_i^{(1)}\Big) \geq \eta^{|\mathcal{U}^{(1)}|}. \quad (116)$$

At the second iteration, we need to bound

$$p^{(2)} \doteq \mathbb{P}\left(\zeta_1^{(2)} \in \bar{\mathcal{I}}_1^{(2)} \wedge \cdots \wedge \zeta_\ell^{(2)} \in \bar{\mathcal{I}}_\ell^{(2)}\right). \quad (117)$$

Using relation (113) we rewrite the previous probability as

$$p^{(2)} = \mathbb{P}\Big(\bigwedge_{i \in \mathcal{U}^{(1)}} \zeta_i^{(2)} \in \bar{\mathcal{I}}_i^{(2)} \wedge \bigwedge_{i \in \mathcal{R}^{(2)}} \zeta_i^{(2)} \in \bar{\mathcal{I}}_i^{(2)}\Big), \quad (118)$$

where we simply grouped the events in two classes, that are such that $\mathcal{U}^{(1)} \cup \mathcal{R}^{(2)} = \{1, \ldots, \ell\}$. The algorithm does not further updates the quantities referred to the indices $i \in \mathcal{U}^{(1)}$ (the intervals that already contain a single integers remain untouched, see lines lines 30–31 of Algorithm 1), and it holds $\bar{\mathcal{I}}_i^{(2)} = \bar{\mathcal{I}}_i^{(1)}$, and $\zeta_i^{(2)} = \zeta_i^{(1)}$ for all $i \in \mathcal{U}^{(1)}$. Therefore, we rewrite (118) as

$$p^{(2)} = \mathbb{P}\Big(\bigwedge_{i \in \mathcal{U}^{(1)}} \zeta_i^{(1)} \in \bar{\mathcal{I}}_i^{(1)} \wedge \bigwedge_{i \in \mathcal{R}^{(2)}} \zeta_i^{(2)} \in \bar{\mathcal{I}}_i^{(2)}\Big). \quad (119)$$

Applying the chain rule, $p^{(2)}$ becomes

$$\begin{aligned}p^{(2)} &= \mathbb{P}\Big(\bigwedge_{i \in \mathcal{R}^{(2)}} \zeta_i^{(2)} \in \bar{\mathcal{I}}_i^{(2)} \,\Big|\, \bigwedge_{i \in \mathcal{U}^{(1)}} \zeta_i^{(1)} \in \bar{\mathcal{I}}_i^{(1)}\Big) \times \\ &\quad \times \mathbb{P}\Big(\bigwedge_{i \in \mathcal{U}^{(1)}} \zeta_i^{(1)} \in \bar{\mathcal{I}}_i^{(1)}\Big). \end{aligned} \quad (120)$$

We can then apply the bound (116):

$$p^{(2)} \geq \eta^{|\mathcal{U}^{(1)}|} \mathbb{P}\Big(\bigwedge_{i \in \mathcal{R}^{(2)}} \zeta_i^{(2)} \in \bar{\mathcal{I}}_i^{(2)} \,\Big|\, \bigwedge_{i \in \mathcal{U}^{(1)}} \zeta_i^{(1)} \in \bar{\mathcal{I}}_i^{(1)}\Big). \quad (121)$$

By definition, all indices $i \in \mathcal{U}^{(1)}$ are such that the confidence interval $\bar{\mathcal{I}}_i^{(1)}$ contains a single integer, say $u_i$, which is the only candidate for $\gamma_i^\circ$. Therefore, the event $\zeta_i^{(1)} \in \bar{\mathcal{I}}_i^{(1)}$ implies $\gamma_i^\circ = u_i$ and we can write

$$p^{(2)} \geq \eta^{|\mathcal{U}^{(1)}|} \mathbb{P}\Big(\bigwedge_{i \in \mathcal{R}^{(2)}} \zeta_i^{(2)} \in \bar{\mathcal{I}}_i^{(2)} \,\Big|\, \bigwedge_{i \in \mathcal{U}^{(1)}} \gamma_i^\circ = u_i\Big). \quad (122)$$

Furthermore, the vector $\boldsymbol{\zeta}_{\mathcal{R}^{(2)}}^{(2)}$ was obtained by conditioning over the values found for the elements in $i \in \mathcal{U}^{(1)}$ (line 38 of Algorithm 1), therefore another conditioning on the same event produces no effect:

$$\begin{aligned}p^{(2)} &\geq \eta^{|\mathcal{U}^{(1)}|} \mathbb{P}\Big(\bigwedge_{i \in \mathcal{R}^{(2)}} \zeta_i^{(2)} \in \bar{\mathcal{I}}_i^{(2)} \,\Big|\, \bigwedge_{i \in \mathcal{U}^{(1)}} \gamma_i^\circ = u_i\Big) \\ &= \eta^{|\mathcal{U}^{(1)}|} \mathbb{P}\Big(\bigwedge_{i \in \mathcal{R}^{(2)}} \zeta_i^{(2)} \in \bar{\mathcal{I}}_i^{(2)}\Big). \end{aligned} \quad (123)$$

Since $\boldsymbol{\zeta}_{\mathcal{R}^{(2)}}^{(2)}$ is again, by construction, a Gaussian random vector with mean $\boldsymbol{\gamma}_{\mathcal{R}^{(2)}}$ and covariance matrix $\boldsymbol{P}_{\mathcal{R}^{(2)}}^{(2)}$ we can apply again Lemma 24 obtaining the bound

$$p^{(2)} \geq \eta^{|\mathcal{U}^{(1)}|} \cdot \eta^{|\mathcal{R}^{(2)}|} = \eta^{|\mathcal{U}^{(1)}|+|\mathcal{R}^{(2)}|}, \quad (124)$$

which becomes $p^{(2)} \geq \eta^\ell = \alpha$, using (114) for $k = 1$. Iterating the same reasoning, at iteration $k$, we can easily obtain the desired bound

$$p^{(k)} \geq \eta^{|\mathcal{U}^{(1)}|+\cdots+|\mathcal{U}^{(k)}|+|\mathcal{R}^{(k+1)}|} = \eta^\ell = \alpha. \quad (125)$$

∎

**Corollary 20** (Correctness of INTEGER-SCREENING). *The integer vector $\boldsymbol{\gamma}^\circ$ is in the set $\Gamma$ returned by Algorithm 1 with probability no smaller than $\alpha$.*

*Proof:* Assume that Algorithm 1 performs $K \leq \ell$ iterations. Since $\Gamma \doteq \Gamma_1^{(K)} \times \ldots \times \Gamma_\ell^{(K)}$ the following equality holds:

$$\mathbb{P}\left(\boldsymbol{\gamma}^\circ \in \Gamma\right) = \mathbb{P}\left(\gamma_1^\circ \in \Gamma_1^{(K)} \wedge \ldots \wedge \gamma_\ell^\circ \in \Gamma_\ell^{(K)}\right). \quad (126)$$

All the integers that are in $\mathcal{I}_i^{(K)}$ are also in $\Gamma_i^{(K)}$, then the previous expression is the same as

$$\mathbb{P}\left(\boldsymbol{\gamma}^\circ \in \Gamma\right) = \mathbb{P}\left(\gamma_1^\circ \in \mathcal{I}_1^{(K)} \wedge \ldots \wedge \gamma_\ell^\circ \in \mathcal{I}_\ell^{(K)}\right). \quad (127)$$

From Proposition 19 it follows that $\mathbb{P}\left(\boldsymbol{\gamma}^\circ \in \Gamma\right) \geq \alpha$. ∎

*C. Optimal choice of the cycle basis matrix*

So far we have not discussed how to choose the cycle basis matrix $\boldsymbol{C}$ and if there is a particular selection that turns out to be more convenient in the screening of admissible vectors $\boldsymbol{\gamma}^\circ$. Note that the choice of $\boldsymbol{C}$ influences the quality of the estimator $\hat{\boldsymbol{\gamma}}$, appearing in both the expression of the estimate and in its covariance $\boldsymbol{P}_\gamma$, as per (103) and (104). Therefore, we are now interested in investigating the choice of $\boldsymbol{C}$ that leads to the most informative estimate of $\boldsymbol{\gamma}^\circ$.

**Algorithm 1**: INTEGER-SCREENING

1  **input**:
2     a real vector $\hat{\boldsymbol{\gamma}} \in \mathbb{R}^\ell$
3     a positive definite matrix $\boldsymbol{P}_\gamma \in \mathbb{R}^{\ell \times \ell}$
4     a confidence level $\alpha \in (0,1)$
5  **precondition**: $\hat{\boldsymbol{\gamma}} \sim \mathcal{N}(\boldsymbol{\gamma}^\circ, \boldsymbol{P}_\gamma)$
6  **output**: a subset of $\mathbb{Z}^\ell$ containing $\boldsymbol{\gamma}^\circ$ with probability at least $\alpha$
7  **variables**:
8     $\mathcal{U}^{(k)} \subseteq \{1, \ldots, \ell\}$  # Indices determined at iter. $k$
9     $\mathcal{R}^{(k)} \subseteq \{1, \ldots, \ell\}$  # Indices that are still ambiguous at iter. $k$
10    $\langle \zeta_{\mathcal{R}^{(k)}}, \boldsymbol{P}_{\mathcal{R}^{(k)}} \rangle$  # Current estimate of $\gamma_i^\circ$, $i \in \mathcal{R}^{(k)}$
11    $\mathcal{I}_i^{(k)} \subset \mathbb{R}$  # Current confidence interval for $\gamma_i^\circ$, $i=\{1,\ldots,\ell\}$
12    $\Gamma_i^{(k)} \subset \mathbb{Z}$  # Current admissible values for $\gamma_i^\circ$, $i=\{1,\ldots,\ell\}$
13
14  $\mathcal{R}^{(1)} \leftarrow \{1, \ldots, \ell\}$
15  $\langle \zeta_{\mathcal{R}^{(1)}}, \boldsymbol{P}_{\mathcal{R}^{(1)}} \rangle \leftarrow \langle \hat{\boldsymbol{\gamma}}, \boldsymbol{P}_\gamma \rangle$
16  **for** $k$ **in** $\{1, 2, \ldots\}$:
17     # Compute confidence sets using marginal probabilities
18     $\mathcal{U}^{(k)} = \emptyset$
19     **for** $i$ **in** $\mathcal{R}^{(k)}$:
20        $\eta = \alpha^{\frac{1}{\ell}}$
21        $b_i \leftarrow \sqrt{\boldsymbol{P}_{ii}^{(k)} \chi_{1,\eta}^2}$, with $\eta = \alpha^{1/\ell}$
22        $\mathcal{I}_i^{(k)} \leftarrow [\zeta_i^{(k)} - b_i, \zeta_i^{(k)} + b]$
23        $\Gamma_i^{(k)} \leftarrow \mathcal{I}_i^{(k)} \cap \mathbb{Z}$
24        # Check if the set contains a single integer
25        **if** $|\Gamma_i^{(k)}| = 1$:
26           $\mathcal{U}^{(k)} \leftarrow \mathcal{U}^{(k)} \cup \{i\}$
27     **end**
28     # Preserve the previous confidence sets
29     **for** $i$ **in** $\{1, \ldots, \ell\} \setminus \mathcal{R}^{(k)}$:
30        $\mathcal{I}_i^{(k)} \leftarrow \mathcal{I}_i^{(k-1)}$
31        $\Gamma_i^{(k)} \leftarrow \Gamma_i^{(k-1)}$
32     **end**
33     # Update the set of indices that are still ambiguous
34     $\mathcal{R}^{(k+1)} \leftarrow \mathcal{R}^{(k)} \setminus \mathcal{U}^{(k)}$
35     **if** $\mathcal{U}^{(k)} = \emptyset$: # Break if there is no progress
36        **break**
37     # Otherwise condition on the elements that we determined
38     $\langle \zeta_{\mathcal{R}^{(k+1)}}, \boldsymbol{P}_{\mathcal{R}^{(k+1)}} \rangle \leftarrow$ CONDITION($\langle \zeta_{\mathcal{R}^{(k)}}, \boldsymbol{P}_{\mathcal{R}^{(k)}} \rangle, \boldsymbol{\gamma}_{\mathcal{U}^{(k)}} = \Gamma_{\mathcal{U}^{(k)}}$)
39  **end**
40  $\Gamma \leftarrow \Gamma_1^{(k)} \times \Gamma_2^{(k)} \times 1 \ldots \Gamma_\ell^{(k)}$
41  **return** $\Gamma$

External functions:
- CONDITION($\langle \zeta, \boldsymbol{P} \rangle, \boldsymbol{\gamma}_i = u_i$) computes the conditional distribution given the constraints that some components are known.

**Proposition 21** (Optimal choice of cycle basis matrix). *Choosing $\boldsymbol{C}$ to be the minimum (uncertainty) cycle basis matrix* MCB *makes $\hat{\boldsymbol{\gamma}}$ a minimum variance unbiased estimator of $\boldsymbol{\gamma}^\circ$ within the class of estimators $\{\hat{\boldsymbol{\gamma}} = \frac{1}{2\pi} \boldsymbol{C}\boldsymbol{\check{\delta}} : \boldsymbol{C} \in \mathcal{C}_{\mathcal{G}}\}$.*

*Proof:* We already know that $\hat{\boldsymbol{\gamma}}$ is unbiased. For an unbiased estimator, the variance is equal to the mean square error. For a Normally distributed estimator, the mean square error is propertional to the trace of the covariance matrix. Therefore, we have to show that choosing $\boldsymbol{C}$ = MCB minimizes $\text{Trace}(\boldsymbol{P}_\gamma) = \frac{1}{4\pi^2} \text{Trace}(\boldsymbol{C}\boldsymbol{P}_\delta\boldsymbol{C}^\mathsf{T})$.

The $t$-th term on the main diagonal of $\boldsymbol{C}\boldsymbol{P}_\delta\boldsymbol{C}^\mathsf{T}$ is $\boldsymbol{c}_t \boldsymbol{P}_\delta \boldsymbol{c}_t^\mathsf{T} = \sum_{(i,j) \in \boldsymbol{c}_t} \sigma_{ij}^2$, where the notation $(i,j) \in \boldsymbol{c}_t$ means "for all edges that belong to the $t$-th cycle". The previous expression coincides with the weight of the cycle $\boldsymbol{c}_t$ under the weight function $w: (i,j) \to \sigma_{ij}^2$.

Therefore, the trace of $\boldsymbol{P}_\gamma = \boldsymbol{C}\boldsymbol{P}_\delta\boldsymbol{C}^\mathsf{T}$ is equal to the sum of the weights of the cycles in the cycle basis, which by definition is $W(\boldsymbol{C}, w)$. Therefore the minimum cycle basis, which minimizes $W(\boldsymbol{C}, w)$, also minimizes $\text{Trace}(\boldsymbol{P}_\gamma)$. ∎

What is the practical advantage of having a minimum variance estimator in our problem? The confidence set $\Gamma$ that we build in Algorithm 1 is directly influenced by the variance of the estimator $\hat{\boldsymbol{\gamma}}$; in Algorithm 1 the diagonal elements of $\boldsymbol{P}_\gamma$ define the width of the confidence intervals, therefore, since the minimum cycle basis matrix minimizes the sum of the diagonal elements of $\boldsymbol{P}_\gamma$ (i.e., its trace), then it also minimizes the widths of the confidence intervals used for the INTEGER-SCREENING algorithm. Therefore, it enables the determination of a small set $\Gamma$ of admissible integer vectors.

## VII. OVERALL ORIENTATION ESTIMATION ALGORITHM: THE MOLE2D ALGORITHM

We can now summarize the findings we presented so far in a single algorithm, that allows computing a (multi-hypothesis) estimate of robot orientations, which we call MOLE2D (*Multi-hypothesis Orientation-from-Lattice Estimation in 2D*). The algorithm computes the set of integers $\Gamma$ using INTEGER-SCREENING, and then for each vector $\boldsymbol{\gamma} \in \Gamma$, it computes the corresponding orientation estimate $\boldsymbol{\check{\theta}}^{\star|\gamma}$. The output is the set of estimates $\Theta = \{\boldsymbol{\check{\theta}}^{\star|\gamma} : \boldsymbol{\gamma} \in \Gamma\}$.

In the section we show that the MOLE2D algorithm solves the two limitations of the maximum likelihood estimator of Section IV (computation efficiency and need of an assessment for the resulting estimate). In particular, in Section VII-A we show that one of the orientation hypotheses returned by the algorithm is "close" to $\boldsymbol{\check{\theta}}^\circ$ (with desired probability); then, in Section VII-B we show that the algorithm includes only worst-case polynomial operations. The MOLE2D algorithm is reported as a pseudocode in Algorithm 2, and described in the following section.

*1) Algorithm 2 overview:* The input to Algorithm 2 is the reduced incidence matrix $\boldsymbol{A}$ (which exhaustively describes the graph), the measurements $\boldsymbol{\check{\delta}}$, their covariance matrix $\boldsymbol{P}_\delta$, and a tuning parameter $\alpha$ which gives the desired confidence level. The output is the set of estimates $\Theta$.

The first step (line 9) is the computation of a cycle basis of the graph. Any cycle basis will do, but Section VI-C tells us that the choice of the cycle basis can be improved if informed by the covariance $\boldsymbol{P}_\delta$, and the best choice is the minimum cycle basis matrix of the graph. Without loss of generality, we assume that the rows of the cycle basis are ordered according to $(\boldsymbol{C}_T; \boldsymbol{C}_L)$, as described in (5).

The next step (lines 11–13) consists in the computation of a set $\Gamma$ which contains $\boldsymbol{\gamma}^\circ$ with confidence $\alpha$ using the INTEGER-SCREENING algorithm, described in Section VI-B.

The "for" loop in line 14 computes, for each integer vector $\boldsymbol{\gamma} \in \Gamma$, the corresponding real-valued estimate $\boldsymbol{\theta}^{\star|\gamma}$ (line 15), and obtain the wrapped estimate $\boldsymbol{\check{\theta}}^{\star|\gamma}$ by applying the exponential map to $\boldsymbol{\theta}^{\star|\gamma}$ (line 16). The collection of wrapped

estimates is then returned in the set $\{\check{\boldsymbol{\theta}}^{\star|\boldsymbol{\gamma}}\}$, which is the output of the algorithm.

---

**Algorithm 2**: MOLE2D

1  **input**:
2    reduced incidence matrix $\boldsymbol{A}$ (topology of the graph)
3    measurements $\check{\boldsymbol{\delta}}$
4    covariance $\boldsymbol{P}_\delta$
5    confidence level $\alpha \in (0, 1)$
6  **output**: set of estimates of $\check{\boldsymbol{\theta}}^\circ$
7
8  # Compute a cycle basis matrix in canonical form
9  $\boldsymbol{C} = (\boldsymbol{C}_T^\mathsf{T}\ \boldsymbol{C}_L^\mathsf{T})^\mathsf{T} \leftarrow$ COMPUTE-CYCLE-BASIS$(\boldsymbol{A}, \boldsymbol{P}_\delta)$
10 # Compute set $\Gamma$, containing admissible vectors for $\boldsymbol{\gamma}^\circ$
11 $\hat{\boldsymbol{\gamma}} \leftarrow \frac{1}{2\pi} \boldsymbol{C} \check{\boldsymbol{\delta}}$
12 $\boldsymbol{P}_\gamma \leftarrow \frac{1}{4\pi^2} \boldsymbol{C} \boldsymbol{P}_\delta \boldsymbol{C}^\mathsf{T}$
13 $\Gamma \leftarrow$ INTEGER-SCREENING$(\hat{\boldsymbol{\gamma}}, \boldsymbol{P}_\gamma, \alpha)$
14 **for** $\boldsymbol{\gamma}$ **in** $\Gamma$:
15   $\boldsymbol{\theta}^{\star|\boldsymbol{\gamma}} = (\boldsymbol{A}\boldsymbol{P}_\delta^{-1}\boldsymbol{A}^\mathsf{T})^{-1}\boldsymbol{A}\boldsymbol{P}_\delta^{-1}\left(\check{\boldsymbol{\delta}} - 2\pi \begin{pmatrix} \boldsymbol{0}_n \\ \boldsymbol{C}_L^{-1} \end{pmatrix} \boldsymbol{\gamma}\right)$
16   $\check{\boldsymbol{\theta}}^{\star|\boldsymbol{\gamma}} = \langle \boldsymbol{\theta}^{\star|\boldsymbol{\gamma}} \rangle_{2\pi}$
17 **end**
18 **return** $\Theta = \{\check{\boldsymbol{\theta}}^{\star|\boldsymbol{\gamma}}\}$.

External functions:
- COMPUTE-CYCLE-BASIS$(\boldsymbol{A}, \boldsymbol{P}_\delta)$ computes a cycle basis for the graph, possibly informed by the covariance matrix $\boldsymbol{P}_\delta$.

---

### A. Assessment of the estimator

This section is aimed at evaluating the quality of our estimator. More precisely, we want to assess how the set of estimates $\{\check{\boldsymbol{\theta}}^{\star|\boldsymbol{\gamma}}\}$, which is the output of Algorithm 2, relates with $\check{\boldsymbol{\theta}}^\circ$. The assessment of the estimator is given in the following proposition.

**Proposition 22.** *Consider the set of estimators $\Theta$ returned by Algorithm 2. Then, with probability no smaller than $\alpha$, one of the estimators in $\Theta$, say $\check{\boldsymbol{\theta}}^{\star|\boldsymbol{\gamma}^j} \in \Theta$, satisfies the following statements:*
1) *the real-valued estimator $\boldsymbol{\theta}^{\star|\boldsymbol{\gamma}^j}$ is Normally distributed with mean $\check{\boldsymbol{\theta}}^\circ + 2\pi\boldsymbol{p}$ and covariance matrix $(\boldsymbol{A}\boldsymbol{P}_\delta^{-1}\boldsymbol{A}^\mathsf{T})^{-1}$;*
2) *the wrapped estimator $\check{\boldsymbol{\theta}}^{\star|\boldsymbol{\gamma}^j} = \langle \boldsymbol{\theta}^{\star|\boldsymbol{\gamma}^j} \rangle_{2\pi}$ is distributed according to a wrapped Gaussian with mean $\check{\boldsymbol{\theta}}^\circ$ and covariance matrix $(\boldsymbol{A}\boldsymbol{P}_\delta^{-1}\boldsymbol{A}^\mathsf{T})^{-1}$;*

*Proof:* These are direct consequences of Lemma 17 and Corollary 20. Lemma 17 assures that the event $\boldsymbol{\gamma}^\circ \in \Gamma$ is the same as the event "at least one real-valued (respectively, wrapped) orientation estimate is distributed according to a Gaussian (respectively, wrapped Gaussian)", and Corollary 20 assures that such event happens with probability $\alpha$. ∎

A straightforward consequence of Proposition 22 is that, in the case $|\Gamma| = 1$, the single estimate contained in $\Theta$ is distributed according to a wrapped Gaussian around $\check{\boldsymbol{\theta}}^\circ$. Therefore, in the case of $|\Gamma| = 1$, we can draw conclusions that are peculiar of linear estimators and are very rare in nonlinear estimation problems (e.g., Normality). In the experimental section we will show that most of the problem instances that constitute a benchmark for state-of-the-art approaches to SLAM satisfy the condition $|\Gamma| = 1$. Therefore, in common problem instances, our multi-hypothesis estimator returns a single *guaranteed* orientation estimate.

### B. Complexity

This section analyzes the worst-case complexity of the operations included in Algorithm 2. The results help assessing the worst-case performance of the algorithm, although they are very conservative in practice, as we will remark at the end of this section and in the experimental analysis. Let us study the complexity of each step of the algorithm.

*Computation of the cycle basis matrix (line 9):* The complexity of computing $\boldsymbol{C}$ heavily depends on the choice of the cycle basis. In the experimental section we will consider four potential choices of the cycle basis: (i) the fundamental cycle basis built from the odometric spanning tree, (ii) the fundamental cycle basis built from the minimum uncertainty spanning tree, (iii) the minimum cycle basis, and (iv) an approximate minimum cycle basis. The fundamental cycle basis built from the odometric spanning tree implies a complexity that is $\mathcal{O}(n\,\ell)$: the odometric spanning path can be considered a given of the problem and the complexity reduces to fill in the matrix $\boldsymbol{C} \in \mathbb{R}^{\ell \times m}$, which has at most $n+1$ nonzero elements in each row. The fundamental cycle basis built from the minimum spanning tree requires the computation of the minimum spanning tree, which amounts to $\mathcal{O}(m+n)$ and the construction of the matrix $\boldsymbol{C}$ ($\mathcal{O}(n\,m)$), therefore the overall cost is $\mathcal{O}(n\,m)$. When using the minimum cycle basis matrix the dominating cost is the actual computation of the cycle basis, which is $\mathcal{O}(m^3)$, and becomes $\mathcal{O}(n^{3+\frac{2}{\nu}})$ for a $(2\nu-1)$-approximate algorithm.

*Computation of the set $\Gamma$ (lines 11–13):* The computation of the estimator $\hat{\boldsymbol{\gamma}}$ requires at most $\ell\,m$ operations, while the covariance matrix requires $\ell^2 m$ operations (exploiting the fact that $\boldsymbol{P}_\delta$ is diagonal). For the INTEGER-SCREENING algorithm, the worst case is when only one index is added to the set of uniquely determined elements at each iteration. Conditioning is an operation that has cubic operation for general matrices ($O(\ell^3)$). Therefore, in the worst case, the complexity is $\mathcal{O}(\ell^4)$ (the algorithm performs $\ell$ conditioning).

*Computation of $\Theta$ (line 18):* This step requires the computation of $\check{\boldsymbol{\theta}}^{\star|\boldsymbol{\gamma}}$ for each $\boldsymbol{\gamma} \in \Gamma$. Let us first evaluate the complexity of computing a single estimate $\check{\boldsymbol{\theta}}^{\star|\boldsymbol{\gamma}}$. The complexity of computing $\check{\boldsymbol{\theta}}^{\star|\boldsymbol{\gamma}}$ from $\boldsymbol{\theta}^{\star|\boldsymbol{\gamma}}$ (i.e., applying the modulus operation) is $\mathcal{O}(n)$. The expression of $\boldsymbol{\theta}^{\star|\boldsymbol{\gamma}}$ is given in (93) and contains the two matrix inverses $\boldsymbol{C}_L^{-1}$ and $(\boldsymbol{A}\boldsymbol{P}_\delta^{-1}\boldsymbol{A}^\mathsf{T})^{-1}$; in practice, one would not perform these matrix inversions, but would rather solve two linear systems. The solution of the first linear system implies a (worst-case) complexity of $\mathcal{O}(\ell^3)$, while the second implies $\mathcal{O}(n^3)$ complexity. Therefore, the worst-case complexity of line 18 amounts to $\mathcal{O}(m^3)$ (recall that $\ell < m$ and $n \leq m$).

Assuming that the cardinality of $\Gamma$ does not grow with problem dimension, the complexity of Algorithm 2 would be $\mathcal{O}(m^4)$. Note that the cardinality of $\Gamma$ depends on the size of the loops and on the measurement uncertainty rather than on the problem dimension: we already observed in the proof of Proposition 21 that the diagonal elements of $\boldsymbol{P}_\delta$ (that determine $\Gamma$) are essentially the sum of the variances of the measurements along each cycle.

We do not refine this bound because this worst-case analysis is a poor indicator of the actual complexity of algorithm, for two main reasons. First, the matrices involved in the various steps of the algorithm are sparse, therefore the computation of $\check{\theta}^{\star|\gamma}$ (line 14) has a complexity that is far below the upper-bound. Second, if we are careful about the choice of the cycle basis matrix from which $\boldsymbol{P}_\gamma$ is computed, the INTEGER-SCREENING algorithm is able to compute a small set $\Gamma$ in few iterations, therefore the average complexity is essentially that of doing one conditioning.

## VIII. EXPERIMENTAL EVALUATION

This section presents an experimental analysis of the proposed approach and its application to pose graph optimization.

Section VIII-A describes the experimental setup.

Section VIII-B discusses the performance of MOLE2D algorithm in the problem of orientation estimation. The objective is to evaluate how important is the choice of the cycle basis matrix in practice, what is the cardinality of the set of candidate vectors $\Gamma$ in real applications (recall that $|\Gamma| = |\Theta|$), and how fast is the algorithm on common problem instances.

Section VIII-C discusses the use of the orientation estimate produced by MOLE2D as the initial guess for iterative techniques for pose graph optimization, such as Toro and g2o. Results show that MOLE2D improves the robustness of such techniques, making them able to produce a good pose estimate also in scenarios with extreme levels of noise, in which they would usually fail.

### A. Benchmark setup

We used three standard datasets:

INTEL  This dataset, acquired at the *Intel Research Lab* in Seattle[2], includes odometry and range-finder data. Relative pose constraints are derived from scan matching. Data processing details are given in previous work [38].

MITb  This dataset was acquired at the MIT *Killian Court*. Data processing details are given in previous work [37].

M3500  This simulated dataset, also known as *Manhattan world*, was created by Olson *et al.* [10].

To test MOLE2D in more challenging scenarios, we obtained other datasets by adding extra Gaussian noise (with standard deviation $\sigma$) to the M3500 orientation measurements. These new datasets are called M3500a ($\sigma = 0.1\,\mathrm{rad}$), M3500b ($\sigma = 0.2\,\mathrm{rad}$), and M3500c ($\sigma = 0.3\,\mathrm{rad}$).

[2]The dataset is provided by Dirk Hähnel and available online [53].

### B. Effect of different cycle bases on orientation estimation and practical computational cost

Here we only consider the orientation measurements in the pose graph and the corresponding covariance matrix. Regarding the MOLE2D algorithm, we chose a confidence level $\alpha = 0.99$. For the computation of the cycle basis matrices, we used Michail's C++ implementation [54]. The rest of the MOLE2D algorithm is instead implemented in Matlab, which makes extremely simple sparse matrix manipulation.

Four cycle bases are considered, listed here from computationally cheap to expensive (Table III):

FCB$_o$  This is the *fundamental cycle basis* built from the *odometric spanning tree*. Call T$_o$ the odometric spanning tree, which is also a spanning path for the graph. Each cycle of FCB(T$_o$) comprises a chord in the graph with respect to T$_o$, say $(i,j)$, and the unique path in T$_o$ from node $i$ to node $j$

FCB$_m$  This is the *fundamental cycle basis* built from the *minimum uncertainty spanning tree*.

MCB$_a$  A $(2\nu - 1)$-*approximation* of the minimum cycle basis is computed using the algorithm proposed by Kavitha *et al.* [55] (in our tests $\nu = 2$).

MCB  The *minimum uncertainty cycle basis*, computed using the method by Mehlhorn and Michail [56].

In the scenarios INTEL, MITb, and M3500, the INTEGER-SCREENING algorithm is able to identify a single possible value for $\gamma^\circ$, regardless the choice of the cycle basis matrix (Table IV, last column). In the scenarios characterized by extreme noise levels (M3500a–c), the choice of the cycle basis truly matters. If one uses the fundamental cycle basis FCB$_o$ the size of $\Gamma$ is too big to be tractable; the explosion of $|\Gamma|$ is partially mitigated by the use of FCB$_m$, that, however, fails to produce a reasonably small number of vectors in $\Gamma$ in the scenario M3500c. Using a minimum cycle basis gives a small cardinality of $\Gamma$, respectively, 1, 3, and 16, for the cases M3500a, M3500b, and M3500c, with no observed difference between the exact minimum cycle basis MCB and the approximation MCB$_a$.

TABLE III
COMPUTATION TIME FOR CYCLE BASIS MATRICES (SECONDS)

|  | $n$ | $m$ | FCB$_o$ | FCB$_m$ | MCB$_a$ | MCB |
|---|---|---|---|---|---|---|
| INTEL | 1228 | 1505 | ≤0.01 | ≤0.01 | 0.09 | 0.20 |
| MITb | 808 | 828 | ≤0.01 | ≤0.01 | 0.01 | 0.01 |
| M3500 | 3500 | 5599 | ≤0.01 | 0.30 | 1.11 | 1.54 |

As predicted by Proposition 21, the minimum cycle bases minimize the number of iterations in the INTEGER-SCREENING (Table IV, second column). Moreover, the minimum cycle bases are able to determine most of the components of $\gamma^\circ$ (e.g., 95%) in the first iteration (Table IV, third column). Finally, the minimum cycle bases require to manage matrices with lower density, when computing matrix inverse (Table IV, fourth column).

All these elements provide a computational advantage when using the minimum cycle bases in the INTEGER-SCREENING (Table V, third column). The minimum cycle basis matrices are usually more sparse, and this also constitute an advantage

TABLE IV
PERFORMANCE OF INTEGER-SCREENING

| | cycle basis | K | | $u$ (%) | $d$ (%) | $|\Gamma|$ |
|---|---|---|---|---|---|---|
| INTEL | FCB$_o$ | 1 | | 100.00 | n/a | 1 |
| | FCB$_m$ | 1 | | 100.00 | n/a | 1 |
| | MCB$_a$ | 1 | | 100.00 | n/a | 1 |
| | MCB | 1 | | 100.00 | n/a | 1 |
| MIT | FCB$_o$ | 2 | iter. 1 | 80.00 | 25.78 | 16 |
| | | | iter. 2 | 20.00 | n/a | 1 |
| | FCB$_m$ | 1 | | 100.00 | n/a | 1 |
| | MCB$_a$ | 1 | | 100.00 | n/a | 1 |
| | MCB | 1 | | 100.00 | n/a | 1 |
| M3500 | FCB$_o$ | 5 | iter. 1 | 52.92 | 1.69 | $>10^{100}$ |
| | | | iter. 2 | 21.54 | 15.61 | $>10^{100}$ |
| | | | iter. 3 | 20.06 | 100.00 | $>10^{50}$ |
| | | | iter. 4 | 5.12 | 100.00 | 972 |
| | | | iter. 5 | 0.36 | n/a | 1 |
| | FCB$_m$ | 2 | iter. 1 | 98.62 | 2.41 | $>10^9$ |
| | | | iter. 2 | 1.38 | n/a | 1 |
| | MCB$_a$ | 2 | iter. 1 | 99.95 | 0.48 | 3 |
| | | | iter. 2 | 0.05 | n/a | 1 |
| | MCB | 2 | iter. 1 | 99.95 | 0.44 | 3 |
| | | | iter. 2 | 0.05 | n/a | 1 |
| M3500a | FCB$_o$ | 6 | | – | – | $>10^{30}$ |
| | FCB$_m$ | 3 | | – | – | 8 |
| | MCB$_a$ | 2 | | – | – | 1 |
| | MCB | 2 | | – | – | 1 |
| M3500b | FCB$_o$ | 29 | | – | – | $>10^{40}$ |
| | FCB$_m$ | 4 | | – | – | 27 |
| | MCB$_a$ | 3 | | – | – | 3 |
| | MCB | 3 | | – | – | 3 |
| M3500c | FCB$_o$ | 9 | | – | – | $>10^{100}$ |
| | FCB$_m$ | 7 | | – | – | $>10^4$ |
| | MCB$_a$ | 4 | | – | – | 16 |
| | MCB | 4 | | – | – | 16 |

This table reports a set of statistics for each scenario and for each choice of the cycle basis matrix:
1) the number of iterations $K$ performed in the INTEGER-SCREENING, reporting details of each iteration when significative;
2) the percentage of elements that are uniquely determined at $k$-th iteration, i.e., $u \doteq |\mathcal{U}^{(k)}|/\ell$ (in percentage);
3) the *density* $d$ of the matrix to be inverted to compute the conditional Gaussian probability in line 38 of INTEGER-SCREENING. The *density* is defined as the number of non-zero elements in the matrix over the total number of elements (in percentage);
4) the number of admissible vectors $\Gamma^{(k)}$ at iteration $k$, and the cardinality of the resulting set $\Gamma$.

The symbol "n/a" denotes that there is no matrix to invert (the algorithm exits the main loop because all elements of $\gamma^\circ$ have been determined). In the cells with "–" we omitted the details of the iterations for brevity.

TABLE V
COMPUTATION TIME FOR MOLE2D (SECONDS)

| | phase | Computation of $\hat{\gamma}$ and $P_\gamma$ | INTEGER-SCREENING | Computation of $\Theta$ from $\Gamma$ | Total |
|---|---|---|---|---|---|
| | lines | 11–12 | 13 | 14–16 | |
| INTEL | FCB$_o$ | 0.07 | 0.04 | $\leq 0.01$ | 0.10 |
| | FCB$_m$ | $\leq 0.01$ | 0.03 | $\leq 0.01$ | 0.04 |
| | MCB$_a$ | $\leq 0.01$ | 0.05 | $\leq 0.01$ | 0.05 |
| | MCB | $\leq 0.01$ | 0.04 | $\leq 0.01$ | 0.04 |
| MIT | FCB$_o$ | $\leq 0.01$ | 0.04 | $\leq 0.01$ | 0.04 |
| | FCB$_m$ | $\leq 0.01$ | 0.03 | $\leq 0.01$ | 0.03 |
| | MCB$_a$ | $\leq 0.01$ | 0.03 | $\leq 0.01$ | 0.03 |
| | MCB | $\leq 0.01$ | 0.03 | $\leq 0.01$ | 0.03 |
| M3500 | FCB$_o$ | 0.72 | 0.79 | $\leq 0.01$ | 1.52 |
| | FCB$_m$ | $\leq 0.01$ | 0.47 | $\leq 0.01$ | 0.47 |
| | MCB$_a$ | $\leq 0.01$ | 0.21 | $\leq 0.01$ | 0.22 |
| | MCB | $\leq 0.01$ | 0.21 | $\leq 0.01$ | 0.22 |
| M3500a | FCB$_o$ | 0.72 | 0.80 | ($\Gamma$ too large to continue) | |
| | FCB$_m$ | $\leq 0.01$ | 0.36 | 0.04 | 0.4 |
| | MCB$_a$ | $\leq 0.01$ | 0.21 | $\leq 0.01$ | 0.22 |
| | MCB | $\leq 0.01$ | 0.21 | $\leq 0.01$ | 0.22 |
| M3500b | FCB$_o$ | 0.71 | 1.19 | ($\Gamma$ too large to continue) | |
| | FCB$_m$ | $\leq 0.01$ | 0.51 | 0.12 | 0.64 |
| | MCB$_a$ | $\leq 0.01$ | 0.23 | 0.03 | 0.26 |
| | MCB | $\leq 0.01$ | 0.23 | 0.03 | 0.26 |
| M3500c | FCB$_o$ | 0.72 | 0.72 | ($\Gamma$ too large to continue) | |
| | FCB$_m$ | $\leq 0.01$ | 0.48 | ($\Gamma$ too large to continue) | |
| | MCB$_a$ | $\leq 0.01$ | 0.23 | 0.15 | 0.38 |
| | MCB | $\leq 0.01$ | 0.23 | 0.14 | 0.37 |

in the computation of $\hat{\gamma}$ and $P_\gamma$ (Table V, second column). Finally, since the minimum cycle bases produce a smaller set of hypotheses in $\Gamma$, they require to solve a smaller number of linear systems for passing from $\Gamma$ to $\Theta$ (Table V, fourth column).

While it is clear that the minimum cycle bases have better performance in MOLE2D, they are usually more expensive to compute (Table III). In conclusion, if the noise is moderate the FCB$_m$ offers a good compromise between performance and computational effort. For extreme noise, the approximate minimum cycle basis matrix MCB$_a$ is a choice that assures a similar performance to the MCB while being cheaper to compute.

*C. Robustness of* MOLE2D-*based pose graph optimization*

In this section we show how to exploit the results of this paper in pose graph optimization. The tested scenarios are the ones of Section VIII-A, but now we also consider the relative position measurements (and the corresponding covariance). The first two columns of Figure 5 report the pose graphs estimated with Toro [11] and g2o [4] (together with the corresponding $\chi^2$ value). It is easy to notice that Toro is more robust to noise than g2o. Generally, gradient methods are known to have a larger basin on convergence [57]. Both methods fail for larger noise as they get stuck in local minima (MITb, M3500a–c). It is interesting that the local minima correspond to incorrect wraparounds in long loops (Figure 5k).

Now we compare these state-of-the-art techniques with a third one that exploits our results: we use the MOLE2D algorithm to compute nodes orientations from relative orientation measurements, and then we substitute this estimate as a first guess for g2o. Note that initial guess of nodes position in MOLE2D+g2o is the odometric one (we only bootstrapped the orientation guess). Following the recommendation of the previous section, we used the approximate minimum cycle basis matrix within the MOLE2D algorithm. If MOLE2D returns more than one hypothesis, we run MOLE2D+g2o for each possible initial guess and choose the one that achieves the minimum.

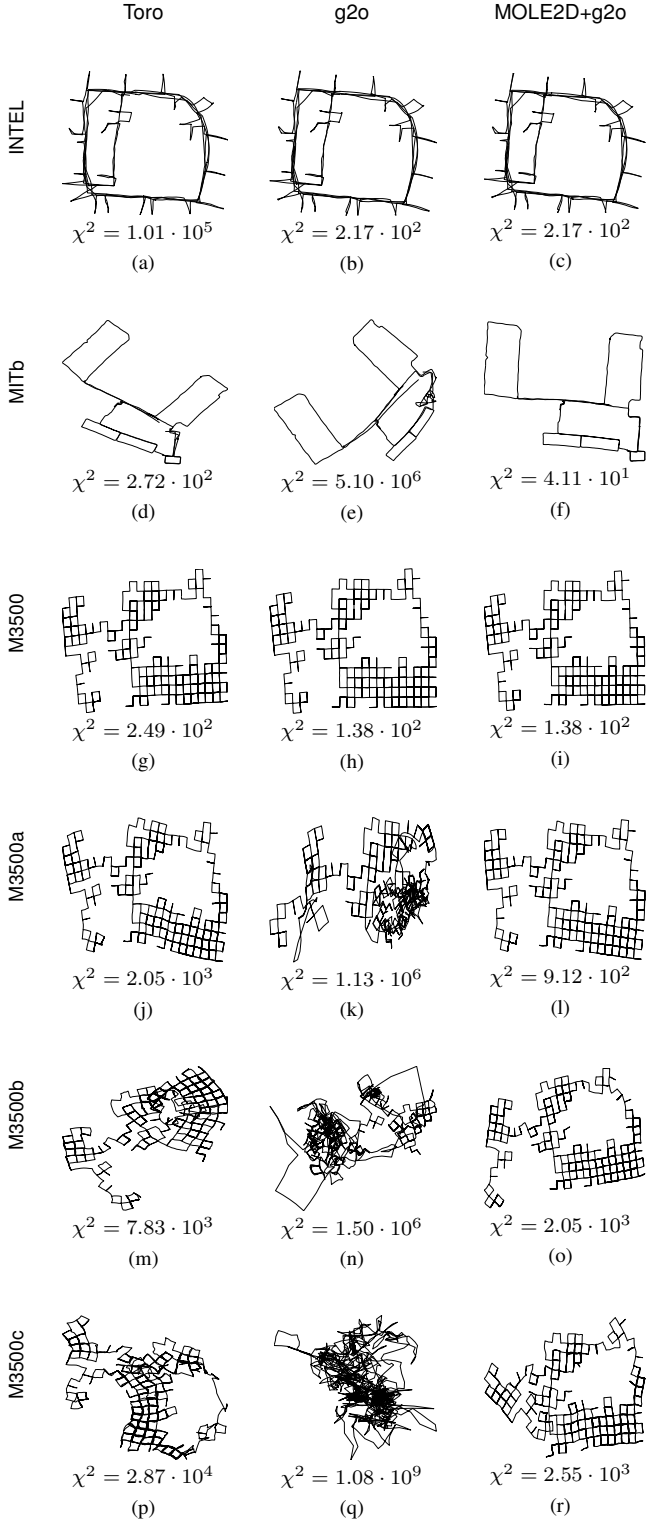

Fig. 5. Estimated pose graphs and corresponding $\chi^2$ values for each benchmarking scenario. The first column reports the results obtained from Toro. The second shows the results obtained from g2o. The third column reports the results obtained by bootstrapping g2o with the orientation estimate of MOLE2D (MOLE2D+g2o).

The third column in Figure 5 shows that this bootstrapping greatly improves the robustness of the iterative solver. In all cases the combination of MOLE2D and g2o attains the smallest observed $\chi^2$ value, and, visually, enable the computation of a correct pose graph estimate.

## IX. CONCLUSION

In this work we discussed the problem of estimating the orientations of nodes in a pose graph from relative orientation measurements. We showed that is possible to recast maximum likelihood orientation estimation in terms of quadratic integer programming. This derivation allowed concluding that the maximum likelihood estimate is almost surely unique and provides a viable solution for computing it without the risk of being trapped in local minima. A deeper consequence of the derivation is that the maximum likelihood orientation estimate does not necessarily lead to an estimate which is "close" to the actual orientation of the nodes. Starting from this observation we devised a multi-hypothesis estimator that enables efficient computation and has guaranteed performance (at least one of the computed estimates is guaranteed to be "close" to the actual orientation of the nodes). We elucidated on the theoretical derivation with some numerical experiments on real and simulated data. As a result, we showed that on common problem instances the multi-hypothesis estimator returns a single estimate. Moreover, a suitable choice of the matrices involved in the estimation enables the computation of a small set of estimates in problems with extreme levels of noise. Finally, we showed that the proposed approach can be used to bootstrap state-of-the-art techniques for pose graph optimization and allows a remarkable boost in their performance, extending their applicability.

Future work includes the analysis of the estimation problem in a 3D setup, which is challenging because SO(3) is not Abelian, so that a nontrival extension of current results is necessary. A second line of research consists in deriving probabilistic guarantees on the *pose* estimate (the results of this paper only guarantee the quality of the *orientation* estimate). Regarding pose graph optimization, a third line of research consists in understanding how the limitations of the maximum likelihood framework manifest their influence (if any) on the estimation of the full poses.

## APPENDIX

**Lemma 23** (Orthogonal projections). *Given a cycle basis matrix $C$ and a reduced incidence matrix $A$ of a connected graph $\mathcal{G}$, for any symmetric positive definite matrix $P$, it holds that*

$$P^{-1}A^{\mathsf{T}}(AP^{-1}A^{\mathsf{T}})^{-1}AP^{-1} + C^{\mathsf{T}}(CPC^{\mathsf{T}})^{-1}C = P^{-1}.$$

*Proof:* Because $P$ is symmetric and positive definite, there exists two symmetric and positive definite matrices $N$ and $M$ such that $M^2 = P$, $N^2 = P^{-1}$, and $N = M^{-1}$.

Following Meyer [58, equation (5.13.3)], the *orthogonal projector* of $NA^{\mathsf{T}}$ is $NA^{\mathsf{T}}(ANNA^{\mathsf{T}})^{-1}AN$ and the orthogonal projector of $MC^{\mathsf{T}}$ is $MC^{\mathsf{T}}(CMMC^{\mathsf{T}})^{-1}CM$.

Because $N$ and $M$ are full rank and $C^\mathsf{T}$ is an orthogonal complement of $A^\mathsf{T}$ (Lemma 1), also $NA^\mathsf{T}$ is an orthogonal complement of $MC^\mathsf{T}$.

Meyer [58, equation (5.13.6)] gives a condition that relates the projectors of two matrices that are orthogonal complements of each other. For our matrices, the relation is

$$NA^\mathsf{T}(ANNA^\mathsf{T})^{-1}AN \\ + MC^\mathsf{T}(CMMC^\mathsf{T})^{-1}CM = \mathbf{I}_m, \tag{128}$$

where $\mathbf{I}_m$ is the identity matrix of size $m$. Pre-multiplying and post-multiplying by $N$, and recalling that $M^2 = P$ and $N^2 = P^{-1}$ we get

$$P^{-1}A^\mathsf{T}(AP^{-1}A^\mathsf{T})^{-1}AP^{-1} \\ + NMC^\mathsf{T}(CPC^\mathsf{T})^{-1}CMN = P^{-1}. \tag{129}$$

By noting that $N = M^{-1}$, we obtain the desired result. ∎

**Lemma 24** (Multiple confidence intervals). *Let $x \in \mathbb{R}^n$ be a Normally distributed random variable with mean $\boldsymbol{\mu}$ and covariance matrix $P$. Given the confidence intervals*

$$\mathcal{I}_i = \left[\mu_i - \sqrt{P_{ii}\chi^2_{1,\eta}}, \ \mu_i + \sqrt{P_{ii}\chi^2_{1,\eta}}\right], \quad i = \{1,\dots,n\},$$

*then*

$$\mathbb{P}(x_1 \in \mathcal{I}_1 \wedge \dots \wedge x_n \in \mathcal{I}_n) \geq \eta^n. \tag{130}$$

*Proof:* The problem of determining multiple confidence sets for possibly correlated random variables has been extensively studied in statistics [59]. In particular, the lemma can be seen as a direct consequence of Theorem 1 in [60]. ∎